\documentclass[acmlarge,screen]{acmart}
\AtBeginDocument{%
  \providecommand\BibTeX{{%
    \normalfont B\kern-0.5em{\scshape i\kern-0.25em b}\kern-0.8em\TeX}}}

\setcopyright{acmcopyright}
\copyrightyear{2018}
\acmYear{2018}
\acmDOI{XXXXXXX.XXXXXXX}

\acmJournal{IMWUT}
\acmVolume{0}
\acmNumber{0}
\acmArticle{111}
\acmMonth{8}

\usepackage{booktabs} % For formal tables
\usepackage[utf8]{inputenc} % Enable linguistic and typographic diversity.
\usepackage{lipsum} % Lorem ipsum dolor sit amet, consectetuer adipiscing elit.
\usepackage{siunitx} % Good SI units typesetting.
\usepackage{multirow} % Fancy table layouts.
\usepackage{tikz,pgfplots} % Plotting
\usepackage{balance} % Balance

\usepackage{amsmath}
\usepackage{array}
\usepackage{graphicx}
\usepackage{wasysym}

\usepackage{xltabular}
\usepackage{threeparttablex}
\usepackage{pdflscape}

\usepackage{booktabs, caption}
\usepackage{longtable}
\usepackage{tabularray}

\usepackage[utf8]{inputenc}
\usepackage[T1]{fontenc}
\usepackage{ltablex}

\usepackage{ctable}

\usepackage{amssymb}
\usepackage{pdfrender}

\usepackage[utf8]{inputenc}
\usepackage{tabularx, makecell, booktabs}

\usepackage{diagbox}
\usepackage{makecell}

\usepackage{pifont} % For bold checkmark
\newcommand{\cmark}{\ding{51}} % Bold checkmark
\newcommand{\xmark}{-} % Placeholder for absent features
\usepackage{geometry} % For adjusting page margins if needed

\usepackage{float}

\begin{document}

\title{Mind Meets Robots: A Review of EEG-Based Brain-Robot Interaction Systems}

\author{Yuchong Zhang}
\orcid{0000-0003-1804-6296}
\email{yuchongz@kth.se}
\affiliation{%
  \institution{KTH Royal Institute of Technology}
  \city{Stockholm}
  \country{Sweden}
}

\author{Nona Rajabi}
\affiliation{%
 \institution{KTH Royal Institute of Technology}
  \city{Stockholm}
  \country{Sweden}}
\email{nonar@kth.se}

\author{Farzaneh Taleb}
\affiliation{%
 \institution{KTH Royal Institute of Technology}
  \city{Stockholm}
  \country{Sweden}}
\email{fatn@kth.se}

\author{Andrii Matviienko}
\affiliation{%
 \institution{KTH Royal Institute of Technology}
  \city{Stockholm}
  \country{Sweden}}
 \email{andriim@kth.se}

\author{Yong Ma}
\affiliation{%
  \institution{University of Bergen}
  \city{Bergen}
  \country{Norway}}
  \email{yong.ma@uib.no}

\author{M\aa{}rten Bj{\"o}rkman}
\affiliation{%
 \institution{KTH Royal Institute of Technology}
  \city{Stockholm}
  \country{Sweden}}
\email{celle@kth.se}

\author{Danica Kragic Jensfelt}
\affiliation{%
 \institution{KTH Royal Institute of Technology}
  \city{Stockholm}
  \country{Sweden}}
\email{dani@kth.se}

\renewcommand{\shortauthors}{Zhang et al.}

\begin{abstract}
Brain-robot interaction (BRI) empowers individuals to control (semi-)automated machines through their brain activity, either passively or actively. In the past decade, BRI systems have achieved remarkable success, predominantly harnessing electroencephalogram (EEG) signals as the central component. This paper offers an up-to-date and exhaustive examination of 87 curated studies published during the last five years (2018-2023), focusing on identifying the research landscape of EEG-based BRI systems. This review aims to consolidate and underscore methodologies, interaction modes, application contexts, system evaluation, existing challenges, and potential avenues for future investigations in this domain. Based on our analysis, we present a BRI system model with three entities: \textit{Brain}, \textit{Robot}, and \textit{Interaction}, depicting the internal relationships of a BRI system. We especially investigate the essence and principles on interaction modes between human brains and robots, a domain that has not yet been identified anywhere. We then discuss these entities with different dimensions encompassed. Within this model, we scrutinize and classify current research, reveal insights, specify challenges, and provide recommendations for future research trajectories in this field. Meanwhile, we envision our findings offer a design space for future human-robot interaction (HRI) research, informing the creation of efficient BRI frameworks.
\end{abstract}

\begin{CCSXML}
<ccs2012>
   <concept>
       <concept_id>10002944.10011122.10002945</concept_id>
       <concept_desc>General and reference~Surveys and overviews</concept_desc>
       <concept_significance>500</concept_significance>
       </concept>
   <concept>
       <concept_id>10003120.10003121.10003128</concept_id>
       <concept_desc>Human-centered computing~Interaction techniques</concept_desc>
       <concept_significance>500</concept_significance>
       </concept>
   <concept>
       <concept_id>10010520.10010553.10010554</concept_id>
       <concept_desc>Computer systems organization~Robotics</concept_desc>
       <concept_significance>300</concept_significance>
       </concept>
 </ccs2012>
\end{CCSXML}

\ccsdesc[500]{General and reference~Surveys and overviews}
\ccsdesc[500]{Human-centered computing~Interaction techniques}
\ccsdesc[300]{Computer systems organization~Robotics}

\keywords{EEG based, brain robot interaction, interaction mode, comprehensive review}

\begin{teaserfigure}
  \includegraphics[width=\textwidth]{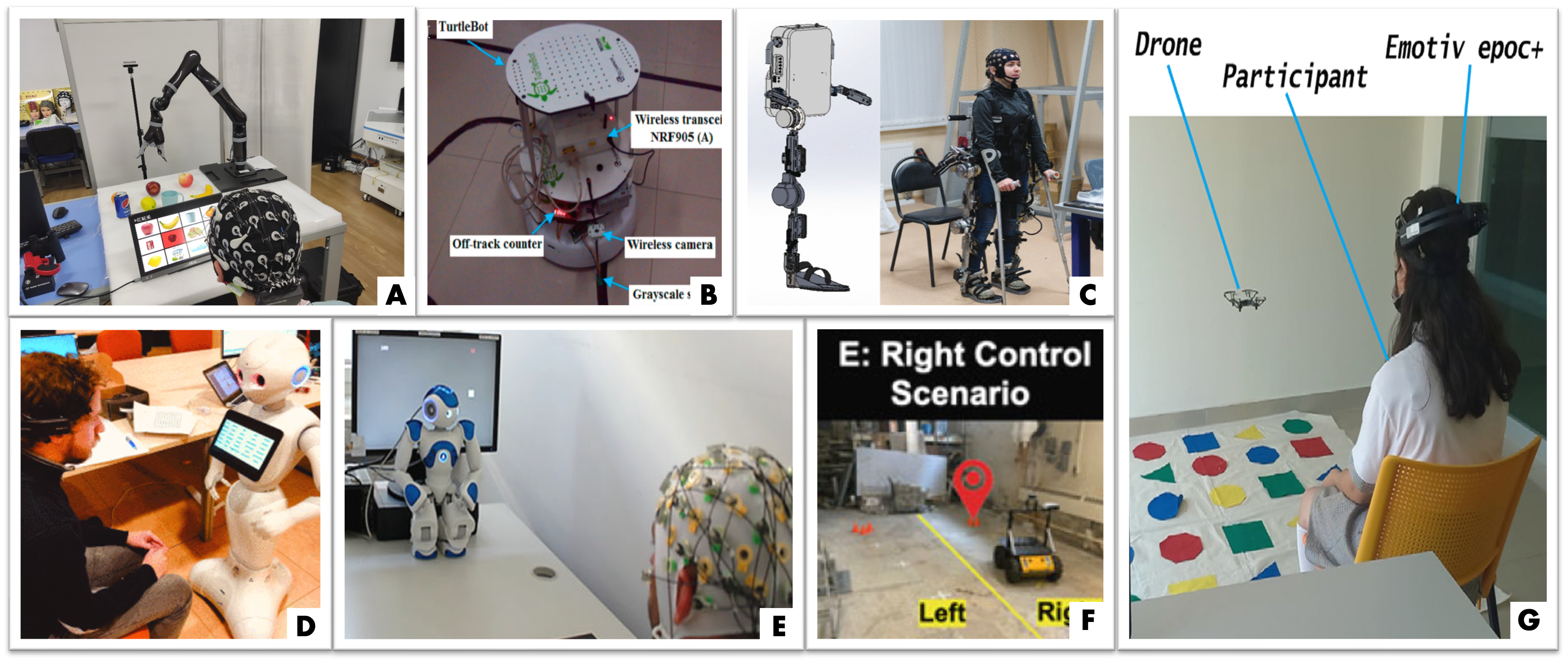}
  \caption{\footnotesize Samples from the reviewed studies in our corpus regarding BRI systems with different robots. (A) Industrial robot \cite{chen2021research} (B) Service robot \cite{wang2018study} (C) Medical robot \cite{ghosh2019continuous} (D) Social robot \cite{staffa2022enhancing} (E) Educational robot \cite{ehrlich2019feasibility} (F) Exploratory robot \cite{liu2021enhanced} (G) Autonomous vehicle \cite{cervantes2023cognidron}.}
  \label{fig:teaser}
\end{teaserfigure}

\received{20 February 2007}
\received[revised]{12 March 2009}
\received[accepted]{5 June 2009}

\maketitle

\section{Introduction}
With the rapidly evolving development of human-computer interaction (HCI) and robotics, the overlapping realm of human-robot interaction (HRI) has attracted significant attention \cite{sheridan2016human}, especially manifested in CHI for the recent years \cite{carros2020exploring,esterwood2021meta,winkle202315}. Nowadays, the inclusion of neuroscience into HCI scenarios has led to an era that transcends traditional boundaries. For instance, the advent of brain-computer interfaces (BCIs) originated in the 1970s \cite{vidal1973toward} has linked the human brain with external intelligent agents by harnessing physiological signals. BCIs are intricate systems capable of circumventing traditional communication pathways to establish direct interaction and control between the human brain and external agents. This is achieved by instantaneously transcribing brain signals from brain activities into operable commands \cite{millan2010combining}. Various approaches exist to measure human biological activities, such as electroencephalogram (EEG), electrooculogram (EOG), electromyogram (EMG), and electrocardiogram (ECG) \cite{perrin2009semi}. Among these, using EEG-based measurements for BCIs and relevant usage has emerged as the predominant approach due to its affordability and exceptional convenience \cite{bi2013eeg}.

Over the past few decades, we have witnessed an evident growth in the development and integration of robots into our daily lives, such as offering assistance in public or at home \cite{bauer2008human,mahdi2022survey}. Robots, as (semi-)automated agents with manifold capabilities, are unexpectedly favored in the industry owing to their exceptional proficiency in movement and operations \cite{goodrich2008human}. As humans and real robots coexist in reality, there arises a necessity for instantaneous mutual understanding, enabling both roles to leverage their unique capabilities and achieve the desired synergy. The integration of robots has significantly enriched the field of HRI, fostering a deeper comprehension and the development of purposeful robotic systems, whether the humans are in close proximity or spanning distances \cite{adams2005human}. Specifically, the dedicated area brain robot interaction (BRI) has gone through remarkable advancement over the last decade, with a discernible and increasingly pronounced trend emerging, particularly in the last five years. Extensive research efforts have been conducted to uncover novel approaches in diverse aspects of BRI to enhance the seamlessness of interaction between human brains and physical robots. The involved robots showcase a wide range of forms and configurations, such as robotic arms \cite{rodriguez1988recursive}, humanoid robots \cite{kajita2014introduction}, and mobile robots \cite{tzafestas2013introduction}. Meanwhile, the collaboration between humans and robots based on complete BRI systems is depicted in various contexts, i.e., guided navigation \cite{chang2021eeg}, knowledge learning \cite{wang2021focus}, and socializing \cite{staffa2022enhancing}.

Notably, the predominant focus within the aforementioned time frame has shown to be on EEG-based BRI which employs nonintrusive technology, ranging from clinical usage to academic research \cite{ibanez2013online,coyle2004suitability}. EEG is a non-invasive technique employed for the assessment of the brain electrical activity, using multiple electrodes attached to the scalp, which offers high temporal resolution but comparatively low spatial resolution.
% Its application extends to the measurement and identification of emotions and intentions. The widespread use is attributed to EEG devices being considered to bear non-intrusiveness, offering ease of use, minimizing harm to the human brain, and being portable. 
As a result, BCIs utilizing EEG signals to capture brain electrical activities have gained widespread popularity \cite{vaughan2003brain}. The unique BRI area has experienced incredible advancements at the intersection of cognitive science, technology, and engineering in the past decade. 
% Over the past few years, the interaction techniques between brains and robots continue to evolve rapidly to provide enhanced comfort to humans and more effective control to robots. As far as we are aware, none of the previous review papers have identified the in-depth aspects of the interactions between humans and robots. 
Thus, more up-to-date literature overview must be formulated on the latest in-depth advancements in EEG-based BRI systems since there were no constructed review articles addressing the research status since 2018, when relevant publications started to thrive unprecedentedly. To address this gap, in this paper, we provide an exhaustive analysis of the current research landscape, encompassing crucial techniques, potential challenges, and prospective research directions for the future development of EEG-based BRI systems. We specifically concentrate more efforts on the analysis of interaction modes and techniques so as to illuminate the future possibilities for the HRI community. The conceptualization of our BRI context is displayed in Fig.~\ref{fig:fw}. Specifically, we intend to answer the following research questions (RQs):

\begin{figure*}[!t]
\centering
  \includegraphics[width=.7\linewidth]{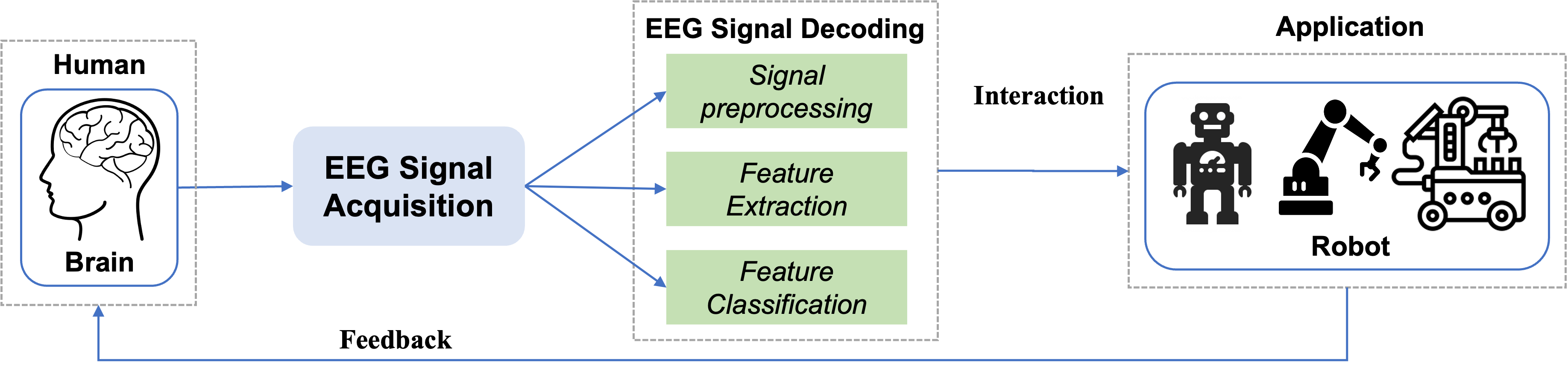}
  \caption{The common framework of a complete BRI system utilizing EEG.}
  \label{fig:fw}
\end{figure*}

\begin{itemize}
    \item \textbf{RQ1}: What are the pivotal techniques, application areas, and evaluation methods used in complete EEG-based BRI systems?
    \item \textbf{RQ2}: What is the nature of the interaction between human brains and robots within the context of EEG?
    \item \textbf{RQ3}: What key insights should be distilled for effective interaction between brain and robot with EEG?
    \item \textbf{RQ4}: What are challenges/limitations and open problems associated with current EEG-based BRI systems?
\end{itemize}

To answer these questions, we examined 87 studies published over the past five years (2018--2023) that explore EEG-based BRI systems.
% In the ensuing sections, we embark on a detailed exploration, delving into the pivotal techniques engaged, diverse applications, existing challenges, and exciting innovations.
Our contributions are manifested in two aspects: (1) providing an overview with in-depth meta-analysis regarding the research landscape of EEG-based BRI (research techniques, application contexts and evaluation methods) as well as guiding the way for future researchers (challenges and outlook), and 2) offering a theoretical contribution in the form of an EEG-based BRI system model and detail the three entities (\textit{Brain}, \textit{Robot}, and \textit{Interaction}) included, with a primary concentration on the interaction between human brains and robots.

% The rest of the paper is elaborated as follows. Section ~\ref{chap:back and re} elucidates the background and related work. In Section ~\ref{chap:method}, we describe the methods used in our review, encompassing our search strategies and data extraction processes. Section ~\ref{chap:model} outlines the devised BRI system model developed based on the meta-analysis of our final corpus with the three entities. For \textit{Brain} entity, the dimensions, namely, EEG signal acquisition, preprocessing, feature extraction and classification are introduced. For \textit{Robot} entity, a diverse range of dimensions entailing robots are summarized. Section ~\ref{chap:interaction} provides an extended analysis of the \textit{Interaction} entity with four dimensions included. In particular, two additional sub-dimensions of Proactive Control are elicited. Section ~\ref{chap:challenges} presents the identified challenges and future directions while Section ~\ref{discussion} shows the discussion of our findings together with the principal limitations. A succinct summary of our paper is encapsulated in Section ~\ref{con}.

\section{Background and Related Work}
\label{chap:back and re}

\subsection{Brain Robot Interaction and EEG Signals}
The term "Brain Robot Interaction" has yet to achieve a universally accepted definition, although it is generally considered a derivative field of HRI \cite{bozinovski2015mental}. This emerging area that establishes a cutting-edge communication bridge between humans and robots particularly through brain signals, holds promise in enhancing the daily lives of individuals with disabilities \cite{zhao2015ssvep,McFarland2008brain}. A typical BRI system operates as a closed-loop control mechanism, integrating human brain signals with contextual feedback. This entails deciphering captured signals from brain activities to formulate commands, thereby instructing the contextual robots to perform desired tasks. Simultaneously, the robot thereby conveys the environmental feedback to the human brain, aiding informed decision-making processes \cite{mao2017progress}. In most BRI systems, human intelligence is highly relied upon to monitor the robot motions based on visual feedback in traditional setups, but machine intelligence has started to flourish and gained significant recognition in recent years \cite{mao2019brain}. Most of the seamless functioning of BRI systems can be attributed to the success of intelligent BCIs integrated with cognitive models tailored for controlling robots \cite{gui2017toward,crawford2015user,chen2016bci}.

\begin{table}[!t]
\centering
\caption{Comparison of contributions between previous review papers and our paper.}
\label{table:contri}
\tiny % Further reduce the font size for space
\setlength{\tabcolsep}{2pt} % Tighten the space between columns
\renewcommand{\arraystretch}{1.2} % Slightly increase space between rows for readability
\begin{tabular}{@{}p{2cm} *{6}{c}@{}}
\toprule
\textbf{Paper} & \textbf{Key Techniques} & \textbf{Robot Categorization} & \textbf{Interaction Mode} & \textbf{Application Contexts} & \textbf{Challenges} & \textbf{Research Outlook} \\
\midrule
Bi et al. 2013 \cite{bi2013eeg} & \cmark & \xmark & \xmark & \cmark & \xmark & \xmark \\
Krishnan et al. 2016 \cite{krishnan2016electroencephalography} & \cmark & \xmark & \xmark & \cmark & \xmark & \xmark \\
Mao et al. 2017 \cite{mao2017progress} & \cmark & \xmark & \xmark & \xmark & \cmark & \cmark \\
Aljalal et al. 2020 \cite{aljalal2020comprehensive} & \cmark & \cmark & \xmark & \xmark & \cmark & \cmark \\
Huang et al. 2021 \cite{huang2021review} & \cmark & \xmark & \xmark & \cmark & \xmark & \cmark \\
Zhang et al. 2023 (Ours) & \cmark & \cmark & \cmark & \cmark & \cmark & \cmark \\
\bottomrule
\label{compa}
\end{tabular}
\end{table}

A complete BRI system collects brain signals as the original input to generate further operations, which can be categorized as invasive and non-invasive. Under invasive context, the brain signals are much stronger while they must be captured inside the brain and always need surgery \cite{aljalal2020comprehensive}. Although non-invasive brain signal acquisition results in weaker outcomes, it merely requires the capture outside the brain with little harm to the human body. In addition, the affordable cost, lower risk, and superior portability comprise other reasons why non-invasive brain signal capturing is preferred in most cases \cite{hwang2013eeg,zhu2010survey,lin2009review}. This exploration adopts the extensive utilization of EEG, a non-invasive neuroimaging technique that captures the electrical activity of the brain \cite{nwagu2023eeg,alimardani2020passive,torres2020eeg}. EEG offers a window into investigating the human mind, enabling the extraction of emotional states and even intentions. Thus, EEG-based BRI systems where the robot/robotic system dominantly employs EEG-based BCIs to interact with humans, have become the most prevailing mechanism in manifold application scenarios \cite{douibi2021toward,li2013hybrid,neuper2003clinical,fan2015step}. Integrating EEG technology with robots paves the way for a new era of interaction between humans and intelligent agents, where neural patterns serve as the central axis of communication. 

\subsection{Current EEG-based BRI Systems}
\begin{figure*}[!t]
\centering
  \includegraphics[width=\linewidth]{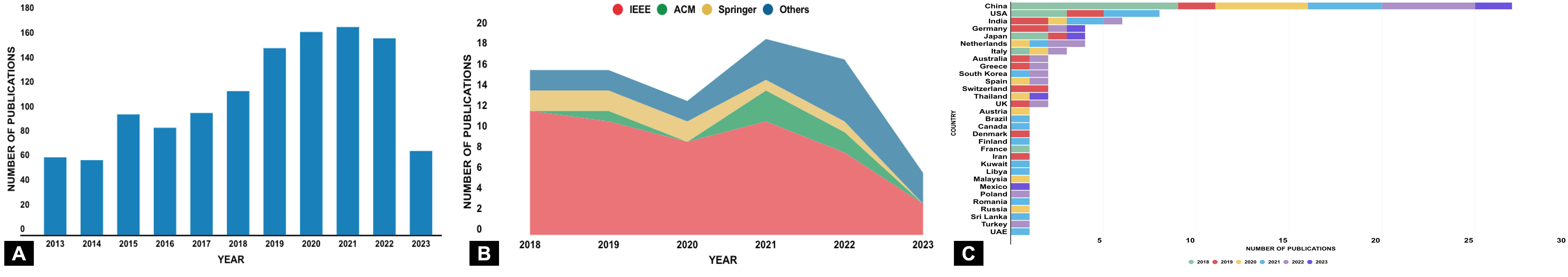}
  \caption{\footnotesize The research agenda of EEG-based BRI in the last few years: A: The number of publications (2013-2023) related to the search results of "EEG-based", "brain", "robot", and "interaction" in the four tested databases. An apparent growing trend is identified especially from 2018. B: The number of papers in our corpus for each year with publishers. For convenience, we listed three main publishers: IEEE, ACM, and Springer, while other publishers are noted as Others. C: The number of studies published for each year with countries.}
  \label{fig:trend}
\end{figure*}

Several reviews have mapped the terrain of EEG-based BRI and associated fields. In 2010, Zhang et al. \cite{zhang2010review} provided a summary of BCI's evolution in industrial robotics and evaluated new commercial BCI products. Subsequently, Si-Mohammed et al. \cite{si2017brain} conducted a literature review merging BCI with augmented reality (AR), covering applications in medicine, robotics, home automation, and brain activity visualization. Hwang et al. \cite{hwang2013eeg} provided a detailed account of EEG-based BCI studies from 2007-2011, and Cao et al. \cite{cao2020review} later updated this landscape in 2020, with a focus on integrating EEG-based BCI and artificial intelligence (AI). In robotics, Bi et al. \cite{bi2013eeg} comprehensively reviewed EEG-driven control for mobile robots, addressing systems, techniques, and evaluations, while Krishhan et al. \cite{krishnan2016electroencephalography} focused on EEG control for assistive robots aiding the disabled and elderly. Aljalal et al. \cite{aljalal2020comprehensive} surveyed EEG signal processing for robot control, highlighting challenges in noninvasive BCI systems. They highlighted encountered challenges of brain-controlled systems at the time. Huang et al. \cite{huang2021review} delved into EEG signal processing methods, spotlighting neural network and deep learning (DL) techniques for signal classification. In 2017, Mao \cite{mao2017progress} reviewed the research output generated from the past years with a precise emphasis on EEG-based BRI interactive systems, which aligns with the thematic intent of our paper the most so far. They identified the key techniques and BCI paradigms, but without focusing on the interaction between brains and robots. In fact, none of the previously mentioned papers investigated the brain-robot interaction mode, a critical subject in the HRI community. Moreover, our observation indicates that research in this area has notably increased since 2018 (Fig. ~\ref{fig:trend}.A), yet reviews synthesizing these recent advances are still missing. While some have then explored EEG-based robot-assisted rehabilitation \cite{berger2019current}, or EEG-based BCI interaction with virtual reality (VR) and AR \cite{nwagu2023eeg}, there's an urgent need for a current review that consolidates the latest EEG-based BRI techniques, highlighting the interaction between human brains and robotic systems. Table ~\ref{compa} showcases the extensive advancements our paper offers over prior similar reviews. We've integrated in-depth insights on key techniques, robot classification, interaction modes, applications, challenges, and future directions.

\section{Method}
\label{chap:method}
% The methodology employed in this review of EEG-based BRI is crucial in ensuring the rigor, inclusiveness, and reliability of the information presented. The following sections outline the approach taken in conducting the review, including the (1) search strategy, (2) the search terms used, (3) filtering, pre-selecting, and manual screening, and (4) the data extraction process.

\subsection{Search Strategy}
\label{search_stra}
\begin{figure*}[!htp]
\centering
  \includegraphics[width=\linewidth]{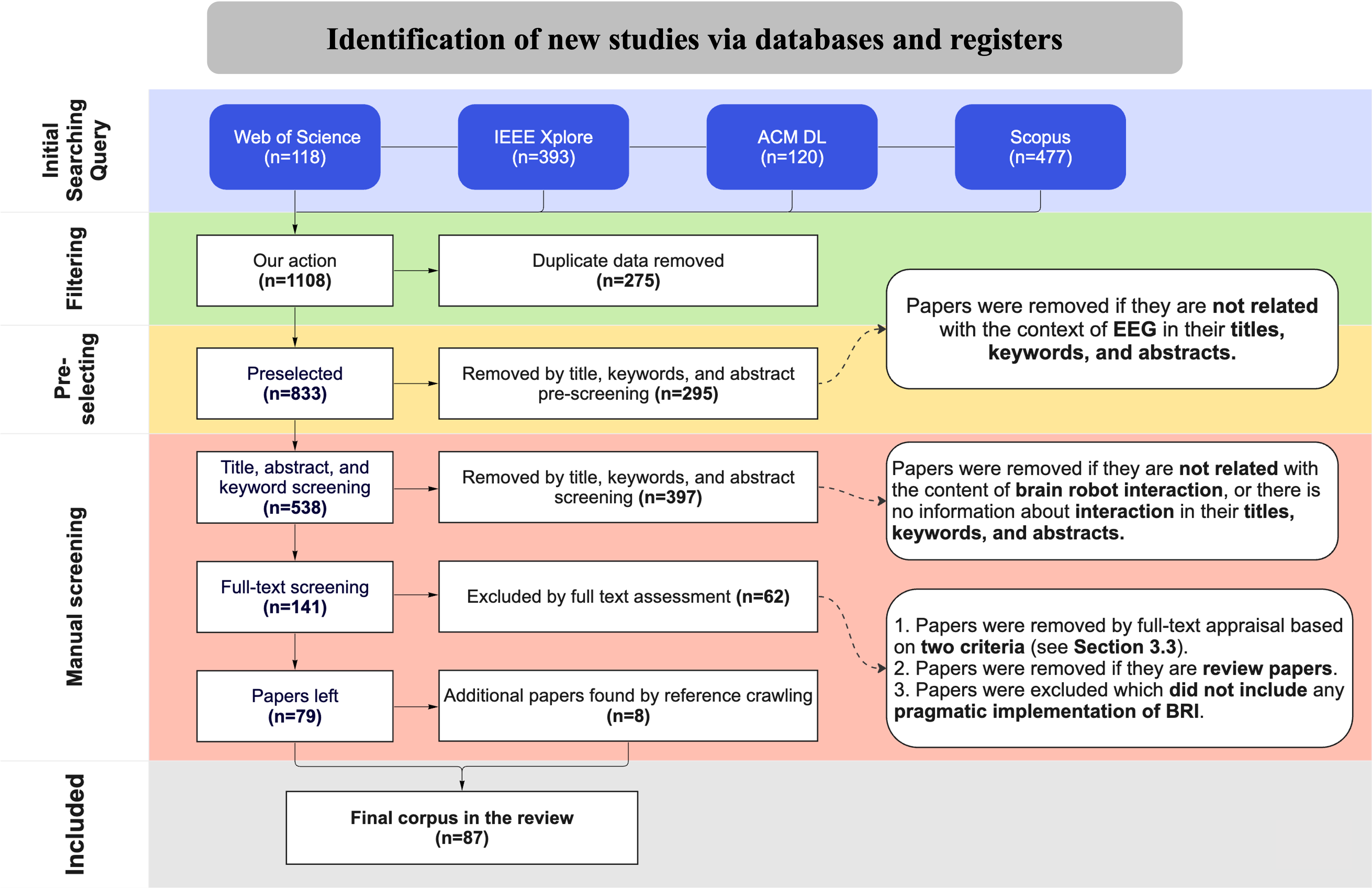}
  \caption{\footnotesize Flow chart of the corpus formulation process with the identification of databases and the initial search query (see Sections ~\ref{search_stra} and ~\ref{search_term}), and filtering, pre-selecting, and the manual screening (see Section ~\ref{filter_screen}), which resulted in 87 full papers.}
  \label{fig:search}
\end{figure*}

The inspiration obtained by methodologies outlined in previously published literature reviews in CHI and other top HCI venues \cite{baytas2019design,pascher2023communicate,van2023systematic,nwagu2023eeg} laid the foundation for our systematic exploration. We adhered to the guidelines set forth by the Preferred Reporting Items for Systematic Reviews (PRISMA) with the updated statement of guidelines \cite{page2021prisma}, as well as the extended framework for scoping reviews \cite{tricco2018prisma}. Four databases, including Web of Science (one of the most comprehensive academic literature databases \cite{chadegani2013comparison}), IEEE Xplore (provides access to more than four million full-text documents from some of the world's most highly cited publications \cite{tomaszewski2021study}), ACM Digital Library (a well-known source incorporating numerous computer science research), and Scopus (a large-scale database containing massive research articles that complement WoS \cite{burnham2006scopus}), were exhaustively scoured, employing a strategic blend of keywords and controlled vocabulary terms. We first established a reference literature dataset by accessing these four databases with the searching the keywords "EEG", "brain", "robot", and "interaction" during the period of 2013--2023. After excluding the repetitions, we obtained 1223 papers in total, which is visualized in Fig.~\ref{fig:trend}.A. As mentioned, we found the publication trend had an obvious increase from 2018 and continued to flourish in the following years. Then, we commenced with the formal literature selection within the scope of our paper (2018--2023). Please refer to Fig.~\ref{fig:search} for an illustrative breakdown of our paper selection process. Five authors were involved in the procedure of paper searching and selection. Elaborations on each distinct stage in this process will be shown in the subsequent sections. During the querying, we incorporated different categories of papers: we unified research articles, conference full papers, and conference proceedings with substantial contributions and advisable pages as full papers, while short papers, posters, late-breaking results as short papers.

\subsection{Search Terms}
\label{search_term}
The formulation of our search strings revolved around the foundational constructs of "EEG," "brain," "robot," and "interaction," along with their complete names and corresponding synonyms. To ensure an expansive coverage reflecting the multifaceted notion of "robot," we extended our purview to encompass related terms such as "exoskeleton" and "wearable" that embody certain robotic attributes. We intentionally avoid using 'brain-computer/machine interface' and 'BCI/BMI' since these queries would lead to incorrect searching results, with massive papers exclusively focusing on BCI after a few attempts. Below, we present the five search strings harnessed in our search procedure, which were devised and crafted by two authors and subsequently endorsed by the entire author group. These strings were strategically employed in the titles, keywords, and abstracts across the array of four databases:

\begin{itemize}
    \item \footnotesize EEG \textbf{OR} EEG-based \textbf{OR} electroencephalogram \textbf{OR} electroencephalogram-based \textbf{AND} robot
    \item \footnotesize EEG \textbf{OR} EEG-based \textbf{OR} electroencephalogram \textbf{OR} electroencephalogram-based \textbf{AND} brain robot interaction \textbf{OR} BRI \textbf{OR} human robot interaction \textbf{OR} HRI
    \item \footnotesize EEG \textbf{OR} EEG-based \textbf{OR} electroencephalogram \textbf{OR} electroencephalogram-based \textbf{AND} exoskeleton \textbf{OR} wearable robots \textbf{OR} wearables
    \item \footnotesize EEG \textbf{OR} EEG-based \textbf{OR} electroencephalogram \textbf{OR} electroencephalogram-based \textbf{AND} brain machine interaction \textbf{OR} brain computer interaction
    \item \footnotesize EEG \textbf{OR} EEG-based \textbf{OR} electroencephalogram \textbf{OR} electroencephalogram-based \textbf{AND} robot interaction \textbf{OR} interactive robot \textbf{OR} robot communication \textbf{OR} communicative robot
\end{itemize}

\subsection{Filtering, Pre-selecting, and Manual Screening}
\label{filter_screen}
This phase began with a specialized process that included filtering duplicates, assessing relevance, and evaluating accessibility within the initial dataset, yielding 1108 papers. After excluding duplicates and removing irrelevant papers based on titles, keywords, and abstracts, we retained 141 papers (Fig. ~\ref{fig:search}). Regarding the full-text screening, each paper from the resultant pool was scrutinized against two fundamental exclusion criteria:

\begin{itemize}
    \item[] (1) Does the paper pertain directly to a fully or predominantly EEG-based context? Consequently, any works that mainly revolved around non-EEG-based scenarios or only tangentially touched upon EEG signals were excluded from our consideration. (e.g. \cite{alimardani2021assessment,liu2020human,kremenski2022brain,stankovic2023human,wang2020humanoid,zhao2020multimodal,li2022preliminary}).
    \item[] (2) Does the paper explicitly detail studies specifically intertwined with human brain robot interaction? This led, for example, to the exclusion of topics that lacked well-defined interaction modalities connecting the human brain and specific robotic systems (e.g. \cite{shi2021novel,xu2022evaluating,nemati2022feature,rodriguez2022affective,rekrut2022improving},), and studies that do not involve the physical, real robotic systems (e.g. \cite{omer2022human,holloman2019leveraging,wang2018exg,tan2021joint,li2020rp,manjunatha2020using}).
\end{itemize}

Review and perspective papers lacking practical BRI implementations, such as \cite{hernandez2020neurophysiological}, were excluded, leaving us with 79 full papers in our final collection (referred to as 'papers' hereafter). Reference crawling added 8 more, totaling 87 papers. Expanding our search to Google Scholar with the same terms yielded no new relevant findings. Our search parameters remained confined to studies published in English within the last five years (1 January 2018 to 31 July 2023), executed by the authors from 8 to 20 August 2023. This compilation encapsulates works emanating from three key publishers, namely IEEE, ACM, and Springer, alongside other publishers (see Fig. ~\ref{fig:trend}.B). Notably, China emerged as a prominent contributor in this field, as depicted in Fig. ~\ref{fig:trend}.C.

\subsection{Data Extraction}
\label{data_extract}
Data extraction involved gathering essential details from each study using a standardized form that included objectives, EEG hardware/software, experimental setup, task paradigms, signal processing techniques, robot types, applications, results, and limitations. Through initial readings of these 87 papers, a series of questions crystallized to facilitate the summarization, consolidation, and comparative analysis of the presented studies:

\begin{itemize}
    \item[] (1) What are the key research questions investigated within the context of the EEG signal?
    \item[] (2) What types of robots/robotic systems are featured in the study? 
    \item[] (3) What are the interaction modes employed between the brain and the robot?
    \item[] (4) Which kind of EEG signals are used and what is the methodology for signal acquisition?
    \item[] (5) What methods are applied in signal decoding?
    \item[] (6) How are these BRI systems assessed, and what metrics are used for evaluation?
    \item[] (7) In what application scenarios are the EEG-based BRI systems explored?
    \item[] (8) What specific tasks are executed within these BRI systems? (Note: In some cases, this aligns with (7).)
    \item[] (9) What findings and outcomes emerge from the research study?
    \item[] (10) How do these outcomes contribute to knowledge acquisition, offering insights, lessons, or guidelines in the field of BRI? (Note: In some papers, this may overlap with (9).)
    \item[] (11) Does the paper present challenges and outline future research directions?
\end{itemize}

\section{The BRI System Model}
\label{chap:model}

\begin{figure*}[!ht]
\centering
  \includegraphics[width=\linewidth]{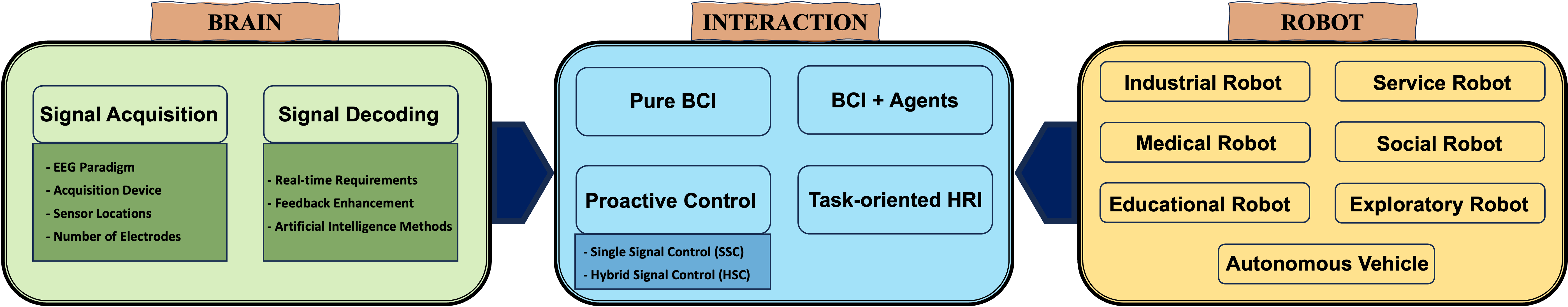}
  \caption{\footnotesize Overview of the BRI system model. The three entities: \textit{Brain}, \textit{Interaction}, and \textit{Robot}, are distilled from our corpus. Human brains and robots are intercorrelated by the \textit{Interaction} entity. Each entity is affiliated with several dimensions, while some of them are comprised of extra sub-dimensions.}
  \label{fig:bri model}
\end{figure*}

Our literature review aimed to deeply understand the brain-robot communication explored in research over five years by analyzing 87 studies. For this, we developed a BRI system model consisting of three key entities (Brain, Robot, and Interaction) as displayed in Fig. ~\ref{fig:bri model}, two of which we discuss in detail in this section, starting with \textit{Brain} and two dimensions Signal Acquisition and Signal Decoding, followed by \textit{Robot} that covers seven dimensions. Additionally, we provide a detailed analysis of application contexts and evaluation methods derived from our corpus. This section tends to address \textbf{RQ} 1. The subsequent section presents the analysis of the third entity of \textit{Interaction}.

\subsection{Brain}
The brain is defined as the source of brain EEG signal extraction and transmission. In the context of a complete EEG-based BRI system, establishing an interactive connection between the human brain and the robot is crucial, which heavily relies on signal (1) acquisition and (2) decoding (Fig. ~\ref{fig:fw}) which we discuss in this part. Table ~\ref{tab:everything} indicates the relevance of this entity within our corpus.

\subsubsection{Signal Acquisition}
\label{acqui}
% The first dimension of the \textit{Brain} entity is a mandatory step for guaranteeing the functioning of an EEG-based BRI system. After the suitable EEG signals used are determined, the acquisition needs to be performed so as to lay the foundation for further use. EEG signals can be collected with appropriate numbers electrodes that are placed on the surface of the head. A wide range of biosignal hardware was employed for EEG data collection in our corpus. The acquisition unit either consisted of an EEG electrode cap linked to an amplifier via wires, or it was an independent EEG device that is typically operated through wireless connections. Upon examination, we observed that none of the prior BRI-related review articles delved into the comprehensive information of the four sub-dimensions we identified. Table ~\ref{tab:brain} displays the distribution of these two distinct acquisition approaches together with the papers that did not specify the signal acquisition processes across our corpus.
The first dimension involves the acquisition of EEG signals, collected using a number of electrodes placed on the scalp by employing a variety of biosignal hardware. The acquisition is typically done through either a wired EEG electrode cap connected to an amplifier or a wireless EEG device. Our review revealed a gap in prior BRI literature, as none thoroughly addressed the four sub-dimensions we identified: EEG paradigms, acquisition devices, sensor locations, and the number of electrodes used across our corpus. Table ~\ref{tab:acquisition} outlines the specific details identified in our corpus, aligning with the sub-dimensions.

\textbf{EEG Paradigms.} Various EEG paradigms have been developed and utilized in research. Following the newest classification proposed by Yadav et al. \cite{yadav4484352eeg}, we identified several paradigms in our corpus: sensory and motor-related, including event-related desynchronization (ERD) and event-related synchronization (ERS), and motor imagery (MI); vision-related, specifically steady-state visually evoked potentials (SSVEP); and cognition-related, encompassing P300 BCI, event-related potential (ERP), and task-based approaches. Our analysis merely reveals one study employing hybrid paradigms while the task-based paradigm is the most prevalent, with MI and SSVEP following in popularity.

\textbf{Acquisition Device.} Numerous bio-signal devices were employed for EEG data collection in our review. These ranged from EEG electrode caps linked to amplifiers to independent EEG units. We identified over 30 distinct EEG acquisition devices across the examined studies and provide an overview of the studies that utilized them underlying wire/wireless connections (~\ref{tab:acquisition}). Particularly, we found that the Emotiv EPOC was the preferred device spanning 19 studies due to its portability and consistent signal quality.

\textbf{Sensor Locations.} The sensors of EEG acquisition devices are typically positioned on various scalp areas, adhering to established standards that correlate electrode placement with the underlying cerebral cortex regions. The most widely recognized standard is the international 10-20 system \cite{jasper1958ten} but we also found instances of the 10-10 system \cite{gordleeva2020real,araujo2021development}. Typically, device sensors are positioned on the scalp to cover key brain regions such as frontal (F), temporal (T), parietal (P), and occipital (O) \cite{cobb1958report}. In certain instances, earlobe areas (A/M) or the midline sagittal plane (Z) are also utilized for grounding or referencing. We categorized sensor locations based on the use brain areas and the inclusion of ground/reference points. As shown in Table ~\ref{tab:acquisition}, we identified six sensor location categories, marking, to our knowledge, the first such classification in EEG device sensor location within review literature. We observed that most studies did not encompass all brain regions with device sensors, while many utilized ground/reference points for accurate signal acquisition.

\begin{landscape}
\begin{table}
\caption{\scriptsize Overview of the Signal Acquisition dimension in \textit{Brain} entity in our corpus. N.S.: not specified.}
\scriptsize\sffamily\centering%
\begin{tabularx}{\linewidth}{>{\hsize=0.5\hsize}X >{\hsize=0.9\hsize}X >{\hsize=0.4\hsize}X >{\hsize=0.3\hsize}X >{\hsize=1.8\hsize}X}
\toprule
\textbf{Sub-dimension} & \textbf{Category} & \textbf{Connection} & \textbf{Papers (\%)} & \textbf{References} \\
\midrule
\multirow{9}{*}{\textbf{EEG Paradigm}} & Sensory and motor-related, SMR/ERD  & --- & 1 (1.1\%) & \cite{nann2020restoring} \\
 & Sensory and motor-related, MRCP  & --- & 1 (1.1\%) & \cite{sugiyama2023eeg} \\
 & Sensory and motor-related, MI & --- & 21 (24.1\%) & \cite{chenga2023using,tang2022wearable,roy2022classification,liu2021enhanced,martinez2020non,penaloza2018android,hernandez2019deep,li2019hybrid,wei2021control,araujo2021development,francis2021eeg,xu2018eeg,boonarchatong2023green,ak2022motor,zhao2020multiple,guo2020nao,gordleeva2020real,kuffuor2018brain,jiang2018brain,wang2018continuous,ai2018cooperative} \\
 & Vision-related, SSVEP & --- & 17 (19.5\%) & \cite{farmaki2022application,chhabra2020bci,fang2022brain,lyu2022coordinating,si2018towards,chu2018robot,chen2021research,abougarair2021implementation,zhang2021mind,du2021vision,aznan2019using,chiuzbaian2019mind,shao2020eeg,karunasena2021single,yuan2018brain,chen2020combination,quiles2022cross} \\
 & Cognition-related, P300 & --- & 8 (9.2\%) & \cite{belkacem2021cooperative,cao2018tactile,braun2019prototype,li2023act,magee2021system,bahman2019robot,rahul2019eeg,ali2021eeg} \\
 & Cognition-related, ERP & --- & 2 (2.2\%) & \cite{aldini2022detection,kompatsiari2018} \\
 & Cognition-related, ErrP & --- & 1 (1.1\%) & \cite{iwane2019inferring} \\
 & Cognition-related, task-based & --- & 33 (37.9\%) & \cite{baka2019talking,prinsen2022passive,wu2022deepbrain,richter2023eeg,memar2018eeg,dissanayake2022eeg,chang2021eeg,aldini2019effect,toichoa2021emotion,staffa2022enhancing,wang2021focus,memar2019objective,lu2022online,alimardani2022robotL,alimardani2020robotM,long2019eeg,yoon2021effect,ogino2018eeg,pawus2022application,cervantes2023cognidron,kilmarx2018sequence,wang2018study,li2022eeg,jo2022eeg,lu2020tractor,roshdy2021machine,sanguantrakul2020development,yu2019design,ghosh2019continuous,mondini2020continuous,rashid2020investigating,korovesis2019robot,ehrlich2019feasibility} \\
 & Cognition-related, IAPS & --- & 1 (1.1\%) & \cite{kim2021affect} \\
 & Hybrid & --- & 1 (1.1\%) & \cite{kar2021eeg} \\
\midrule
\multirow{30}{*}{\textbf{Acquisition Device}} & ADS1299 & wired/wireless & 2 (2.2\%) & \cite{li2022eeg,sanguantrakul2020development} \\
 & BIOPAC MP150  & wired & 1 (1.1\%) & \cite{ghosh2019continuous} \\
 & BioRadio system & wireless & 1 (1.1\%) & \cite{karunasena2021single} \\
 & Biosemi ActiveTwo & wired & 3 (3.4\%) & \cite{tang2022wearable,long2019eeg,lyu2022coordinating} \\
 & BrainProducts MOVE & wireless& 1 (1.1\%) & \cite{aldini2022detection} \\
 & BrainProducts ActiChamp  & wired & 1 (1.1\%) & \cite{ehrlich2019feasibility} \\
 & BrainProducts ActiCap & wired/wireless & 4 (4.6\%) & \cite{martinez2020non,kompatsiari2018,mondini2020continuous,araujo2021development} \\
 & BrainProducts LiveAmp & wireless & 1 (1.1\%) & \cite{nann2020restoring} \\
 & BrainProducts others & wired & 2 (2.2\%) & \cite{kim2021affect,aldini2019effect} \\
 & Cognionics HD-72 & wireless &  1 (1.1\%) & \cite{qian2018affective} \\
 & Cognionics Quick-20  & wireless & 1 (1.1\%) & \cite{aznan2019using} \\
 & EGI system & wired & 1 (1.1\%) & \cite{alimardani2022robotL} \\
 & Emotiv EPOC & wireless & 19 (21.8\%) & \cite{roshdy2021machine,pawus2022application,chang2021eeg,kilmarx2018sequence,yu2019design,bahman2019robot,chu2018robot,chiuzbaian2019mind,chhabra2020bci,staffa2022enhancing,wang2018continuous,ai2018cooperative,kuffuor2018brain,lu2020tractor,shao2020eeg,braun2019prototype,li2023act,ali2021eeg,toichoa2021emotion} \\
 & g.HIAMP & wired & 1 (1.1\%) & \cite{richter2023eeg} \\
 & g.MOBIlab+ & wireless & 1 (1.1\%) & \cite{dissanayake2022eeg} \\
 & g.Tec Nautilus & wired & 1 (1.1\%) & \cite{hernandez2019deep} \\
 & g.USBamp & wired/N.S. & 3 (3.4\%) & \cite{yoon2021effect,alimardani2020robotM,si2018towards} \\
 & Neuracle & wired/N.S. & 3 (3.4\%) & \cite{chen2020combination,abougarair2021implementation,chen2021research} \\
 & Neuroelectrics  & wireless & 1 (1.1\%) & \cite{quiles2022cross} \\
 & NeuroScan SynAmps & wired & 2 (2.2\%) & \cite{fang2022brain,jo2022eeg} \\
 & NeuroScan NuAmps & wireless & 4 (4.6\%) & \cite{li2019hybrid,baka2019talking,guo2020nao,yuan2018brain} \\
 & NeuroScan Others & wired/wireless/N.S. & 4 (4.6\%) & \cite{magee2021system,du2021vision,cao2018tactile,wang2018study} \\
 & NeuroSky Mindwave & wireless/N.S. & 6 (6.9\%) & \cite{rahul2019eeg,rashid2020investigating,wang2021focus,ogino2018eeg,ak2022motor,boonarchatong2023green} \\
 & NeuroSky TGAM & wireless & 1 (1.1\%) & \cite{lu2022online} \\
 & Nexus 32 & wireless & 1 (1.1\%) & \cite{cervantes2023cognidron} \\
 & Nihon Kohden & N.S. & 1 (1.1\%) & \cite{kar2021eeg} \\
 & NVX 52 & wired & 1 (1.1\%) & \cite{gordleeva2020real} \\
 & OpenBCI Cyton & wireless & 1 (1.1\%) & \cite{farmaki2022application} \\
 & OpenBCI Ganglion & N.S. & 1 (1.1\%) & \cite{korovesis2019robot} \\
 & Unicorn Hybrid Black & wireless & 2 (2.2\%) & \cite{belkacem2021cooperative,prinsen2022passive} \\
\midrule
\multirow{6}{*}{\textbf{Sensor Locations}} & One of F,T,P,O Involved & --- & 14 (16.1\%) & \cite{guo2020nao,kar2021eeg,belkacem2021cooperative,farmaki2022application,chiuzbaian2019mind,memar2018eeg,ghosh2019continuous,wang2021focus,kuffuor2018brain,korovesis2019robot,chenga2023using,li2023act,sanguantrakul2020development,martinez2020non} \\
 & Two of F,T,P,O Involved & --- & 21 (24.1\%) & \cite{abougarair2021implementation,araujo2021development,karunasena2021single,zhang2021mind,yuan2018brain,iwane2019inferring,cao2018tactile,chen2021research,bahman2019robot,jiang2018brain,rashid2020investigating,ogino2018eeg,dissanayake2022eeg,yu2019design,du2021vision,chen2020combination,si2018towards,kompatsiari2018,shao2020eeg,lu2022online,rahul2019eeg} \\
 & Three of F,T,P,O Involved & --- & 9 (10.3\%) & \cite{wei2021control,prinsen2022passive,nann2020restoring,fang2022brain,chu2018robot,aznan2019using,quiles2022cross,magee2021system,penaloza2018android} \\
 & All of F,T,P,O Involved & --- & 13 (14.9\%) & \cite{baka2019talking,cervantes2023cognidron,ak2022motor,ehrlich2019feasibility,zhao2020multiple,chang2021eeg,tang2022wearable,wang2018continuous,ai2018cooperative,long2019eeg,staffa2022enhancing,chhabra2020bci,alimardani2022robotL} \\
 & Earlobe (A/M) Involved & --- & 21 (24.1\%) & \cite{cervantes2023cognidron,ak2022motor,zhao2020multiple,wang2018study,wang2018continuous,ai2018cooperative,ogino2018eeg,long2019eeg,staffa2022enhancing,dissanayake2022eeg,chu2018robot,yu2019design,aznan2019using,du2021vision,chhabra2020bci,wang2021focus,kuffuor2018brain,korovesis2019robot,kompatsiari2018,rahul2019eeg,martinez2020non} \\
 & Midline Sagittal Plane (Z) Involved & --- & 22 (25.3\%) & \cite{cervantes2023cognidron,araujo2021development,yoon2021effect,nann2020restoring,karunasena2021single,zhang2021mind,jiang2018brain,wang2018continuous,rashid2020investigating,ai2018cooperative,li2019hybrid,staffa2022enhancing,dissanayake2022eeg,aznan2019using,chen2020combination,korovesis2019robot,kompatsiari2018,quiles2022cross,shao2020eeg,rahul2019eeg,martinez2020non,penaloza2018android} \\
\midrule
\multirow{5}{*}{\textbf{Number of Electrodes}} & No. $\le$ 10 & --- & 37 (42.5\%) & \cite{wang2021focus,xu2018eeg,rashid2020investigating,ogino2018eeg,chiuzbaian2019mind,rahul2019eeg,wu2022deepbrain,roy2022classification,karunasena2021single,guo2020nao,zhang2021mind,li2019hybrid,lu2022online,yuan2018brain,ghosh2019continuous,yu2019design,korovesis2019robot,sanguantrakul2020development,ali2021eeg,bahman2019robot,iwane2019inferring,si2018towards,magee2021system,gordleeva2020real,kar2021eeg,prinsen2022passive,nann2020restoring,fang2022brain,belkacem2021cooperative,dissanayake2022eeg,du2021vision,quiles2022cross,hernandez2019deep,abougarair2021implementation,chen2021research,chen2020combination,kuffuor2018brain} \\
 & 10 $<$ No. $\le$ 20 & --- & 23 (26.4\%) & \cite{alimardani2022robotL,lu2020tractor,roshdy2021machine,braun2019prototype,toichoa2021emotion,alimardani2020robotM,kilmarx2018sequence,yoon2021effect,chang2021eeg,cao2018tactile,staffa2022enhancing,chu2018robot,chhabra2020bci,li2023act,cervantes2023cognidron,wei2021control,araujo2021development,wang2018continuous,ai2018cooperative,jiang2018brain,sugiyama2023eeg,memar2019objective,aznan2019using} \\
 & 20 $<$ No. $\le$ 32 & --- & 14 (16.1\%) & \cite{pawus2022application,ak2022motor,penaloza2018android,wang2018study,kim2021affect,aldini2019effect,jo2022eeg,lyu2022coordinating,aldini2022detection,liu2021enhanced,qian2018affective,ehrlich2019feasibility,long2019eeg,shao2020eeg} \\
 & 32 $<$ No. $\le$ 64 & --- & 5 (5.7\%) & \cite{baka2019talking,martinez2020non,richter2023eeg,tang2022wearable,kompatsiari2018} \\
 & No. $>$ 64 & --- & 3 (3.4\%) & \cite{mondini2020continuous,farmaki2022application,francis2021eeg} \\
\bottomrule
\end{tabularx}
\label{tab:acquisition}
\end{table}

\end{landscape}

\textbf{Number of Electrodes.} We aimed to determine the specific number of electrodes utilized in the signal acquisition, however, several studies did not provide precise details. According to \cite{montoya2021effect}, 64 electrodes are typically adopted in practical cases, however, 20 and 32 electrodes are observed to yield distinctively desired results in subject-independent cases. Since our observation revealed that many studies employ fewer than 10 electrodes, leading to the classification into five categories based on electrode count: 10, 20, 32, and 64 (Table ~\ref{tab:acquisition}). The majority of studies prefer fewer electrodes, with 37 studies using less than 10, and only 8 studies utilizing more than 32 electrodes. %Our compilation aims to aid future researchers in selecting an appropriate number of electrodes for designing targeted BRI systems. 

% \textit{Wired:} Within this sub-dimension, electrodes that are placed on the head are typically linked to an EEG amplifier or recording system through wired connections (i.e., \cite{jo2022eeg,martinez2020non,zhang2021mind}). Wired EEG configurations are in relation to physical connections, being directly attached to the scalp via cables.

% \textit{Wireless:} This sub-dimension pertains to EEG signal acquisition equipment designed as a self-contained system, featuring wireless connectivity options like Bluetooth or WiFi (i.e., \cite{wang2018study,aldini2022detection,lu2020tractor}). Most of the reviewed studies favoured wireless acquisition to transmit signals to eliminate the need for physical tethers. The choice between wired and wireless contexts may result in difference in mobility, comfort, and user experience (UX) during EEG data collection, which is dependent on specific cases.

% \textit{Not specified:} The ways of the connection between the acquisition device and the computer regarding acquiring EEG signals are not specified in the paper (i.e., \cite{ak2022motor,zhao2020multiple,korovesis2019robot}).

\subsubsection{Signal Decoding}
\label{decoding}
The traditional decoding process involves several key steps: preprocessing to remove any artifacts from the data, feature extraction to capture relevant signal information, and feature classification where labels are assigned to features based on predefined classes. Common techniques like band-pass \cite{aldini2019effect,staffa2022enhancing,kompatsiari2018} and notch filters \cite{chu2018robot,roy2022classification,zhang2021mind} for preprocessing, and Canonical Correlation Analysis (CCA) \cite{chen2021research,zhang2021mind,du2021vision} and Common Spatial Patterns (CSP) \cite{wei2021control, araujo2021development} for feature extraction and classification, are well-documented in prior literature. However, the recent surge in AI like DL related technologies has introduced novel advancements in this area. Besides, the real-time requirements and feedback enhancement of the decoding process have not yet been investigated in previous reviews. Table ~\ref{tab:decoding} displays the studies identified alongside the three sub-dimensions: real-time requirement, feedback enhancement and AI methods.

\textbf{Real-time Requirements.} In the majority of the studies reviewed, EEG signals were processed in real-time, meaning the brain's electrical activity was analyzed instantly or with negligible delay, facilitating immediate device interaction or interpretation. Our corpus provided a rich source of information, from which we distilled five types of requirements, highlighting the critical factors that contribute to the successful implementation of EEG-based BRI systems.

\begin{itemize}
    \item Low Latency: To maintain a natural and intuitive user experience (UX), the latency from signal acquisition to action or feedback should be minimal, ranging from a few to several hundred milliseconds. This requirement emerged as the second most prevalent in our corpus, as 23 studies involved it.
    \item High Accuracy: Decoding algorithms need to precisely translate EEG signals into accurate commands or responses, reducing errors and misunderstandings for effective interaction. This is crucial for applications where incorrect interpretations can lead to potentially dangerous outcomes (n=15).
    \item High Temporal Resolution: The EEG system requires a high sampling rate to accurately record the brain's fast-changing activity, essential for effective real-time signal decoding. Only a few studies successfully met this criterion (n=8).
    \item Seamless Feedback: Immediate and intuitive feedback from EEG signals to users is crucial for practical applications, especially for neurofeedback and BCI used cases \cite{ogino2018eeg,long2019eeg}, so as to enhance user controllability. The understandable is normally facilitated by preprocessing, feature extraction, and classification. This requirement was present in nearly half of the studies (n=42).
    \item Computational Efficiency: Decoding algorithms should be computationally efficient for real-time processing, often necessitating optimized software and specialized hardware. Notably, only four studies fulfilled this requirement.
\end{itemize}

\begin{table}[!htb]
\caption{Overview of the Signal Decoding dimension in the \textit{Brain} entity.}
\scriptsize\sffamily\centering%
\begin{tabularx}{\linewidth}{>{\hsize=0.75\hsize}X >{\hsize=0.9\hsize}X >{\hsize=0.5\hsize}X >{\hsize=1.7\hsize}X}
\toprule
\textbf{Sub-dimension} & \textbf{Category} & \textbf{Papers (\%)} & \textbf{References} \\
\midrule
\multirow{5}{*}{\textbf{Real-time Requirements}} & Low Latency & 23 (26.4\%) & \cite{belkacem2021cooperative,chhabra2020bci,wu2022deepbrain,sugiyama2023eeg,toichoa2021emotion,tang2022wearable,liu2021enhanced,mondini2020continuous,xu2018eeg,korovesis2019robot,karunasena2021single,chen2020combination,wang2018continuous,jo2022eeg,aldini2022detection,rashid2020investigating,guo2020nao,zhao2020multiple,ehrlich2019feasibility,prinsen2022passive,roshdy2021machine,bahman2019robot,francis2021eeg} \\
 & High Accuracy & 15 (17.2\%) & \cite{rahul2019eeg,wang2018study,cao2018tactile,martinez2020non,magee2021system,chen2021research,roy2022classification,jo2022eeg,aldini2022detection,rashid2020investigating,guo2020nao,francis2021eeg,abougarair2021implementation,penaloza2018android,nann2020restoring} \\
 & High Temporal Resolution & 8 (9.2\%) & \cite{baka2019talking,pawus2022application,aznan2019using,kim2021affect,farmaki2022application,li2019hybrid,zhao2020multiple,ehrlich2019feasibility} \\
 & Seamless Feedback & 42 (48.3\%) & \cite{araujo2021development,chen2021research,roy2022classification,kim2021affect,farmaki2022application,li2019hybrid,prinsen2022passive,roshdy2021machine,bahman2019robot,braun2019prototype,li2023act,qian2018affective,dissanayake2022eeg,chang2021eeg,aldini2019effect,staffa2022enhancing,iwane2019inferring,kompatsiari2018,long2019eeg,yoon2021effect,ogino2018eeg,cervantes2023cognidron,kar2021eeg,chu2018robot,kilmarx2018sequence,li2022eeg,lu2020tractor,zhang2021mind,du2021vision,sanguantrakul2020development,hernandez2019deep,chiuzbaian2019mind,ghosh2019continuous,wei2021control,ali2021eeg,gordleeva2020real,kuffuor2018brain,memar2018eeg,richter2023eeg,abougarair2021implementation,penaloza2018android,nann2020restoring} \\
 & Computational Efficiency & 4 (4.6\%) & \cite{lu2022online,rahul2019eeg,araujo2021development,richter2023eeg} \\
\midrule
\multirow{6}{*}{\textbf{Feedback Enhancement}} & Visual Feedback & 28 (32.2\%) & \cite{baka2019talking, belkacem2021cooperative,farmaki2022application,sugiyama2023eeg,kompatsiari2018,si2018towards,chenga2023using,kilmarx2018sequence,wang2018study,cao2018tactile,abougarair2021implementation,du2021vision,magee2021system,penaloza2018android,yu2019design,chiuzbaian2019mind,bahman2019robot,ghosh2019continuous,mondini2020continuous,rahul2019eeg,yuan2018brain,quiles2022cross,prinsen2022passive,aznan2019using,hernandez2019deep,zhang2021mind,ehrlich2019feasibility,alimardani2022robotL} \\
 & Auditory Feedback & 3 (3.4\%) & \cite{kar2021eeg, richter2023eeg, wang2021focus} \\
 & Multimodal Feedback & 7 (8\%) & \cite{braun2019prototype, li2023act, chen2021research, li2022eeg, roy2022classification,li2019hybrid,dissanayake2022eeg} \\
 & Direct Feedback & 22 (25.3\%) & \cite{prinsen2022passive,aznan2019using,hernandez2019deep,zhang2021mind,ehrlich2019feasibility,alimardani2022robotL,chhabra2020bci,fang2022brain,memar2018eeg,lu2022online,pawus2022application,jo2022eeg,aldini2022detection,martinez2020non,sanguantrakul2020development,wei2021control,araujo2021development,francis2021eeg,rashid2020investigating,staffa2022enhancing,cervantes2023cognidron,kar2021eeg} \\
 & Performance Feedback & 18 (20.1\%) & \cite{cervantes2023cognidron, guo2020nao, qian2018affective, wu2022deepbrain, iwane2019inferring,memar2019objective,long2019eeg,liu2021enhanced,lu2020tractor,shao2020eeg,boonarchatong2023green, kuffuor2018brain, jiang2018brain, chen2020combination, wang2018continuous, ai2018cooperative,lyu2022coordinating,aldini2019effect} \\
 & Neurofeedback & 22 (25.3\%) & \cite{staffa2022enhancing,kim2021affect,dissanayake2022eeg, chang2021eeg,toichoa2021emotion, alimardani2020robotM,yoon2021effect,tang2022wearable,ogino2018eeg,chu2018robot,roshdy2021machine,ali2021eeg,xu2018eeg,ak2022motor,zhao2020multiple,gordleeva2020real,nann2020restoring,korovesis2019robot,karunasena2021single,guo2020nao,lyu2022coordinating,aldini2019effect,richter2023eeg} \\
\midrule
\multirow{3}{=}{\textbf{Artificial Intelligence Methods}} & Traditional Machine Learning (ML) & 42 (48.3\%) & \cite{araujo2021development,cao2018tactile,magee2021system,chen2021research,roy2022classification,farmaki2022application,li2019hybrid,belkacem2021cooperative,chhabra2020bci,xu2018eeg,chen2020combination,wang2018continuous,jo2022eeg,aldini2022detection,rashid2020investigating,guo2020nao,zhao2020multiple,ehrlich2019feasibility,francis2021eeg,braun2019prototype,qian2018affective,chang2021eeg,iwane2019inferring,long2019eeg,ogino2018eeg,kar2021eeg,chu2018robot,lu2020tractor,zhang2021mind,du2021vision,wei2021control,gordleeva2020real,kuffuor2018brain,penaloza2018android,lyu2022coordinating,memar2019objective,si2018towards,shao2020eeg,boonarchatong2023green,yuan2018brain,jiang2018brain,quiles2022cross} \\
 & Deep Learning (DL) & 19 (21.8\%) & \cite{rahul2019eeg,magee2021system,pawus2022application,aznan2019using,li2019hybrid,wu2022deepbrain,tang2022wearable,korovesis2019robot,aldini2022detection,guo2020nao,roshdy2021machine,li2023act,staffa2022enhancing,kar2021eeg,lu2020tractor,sanguantrakul2020development,ghosh2019continuous,fang2022brain,chenga2023using} \\
 & Non-ML/DL & 7 (8\%) & \cite{baka2019talking,wang2018study,yoon2021effect,li2022eeg,hernandez2019deep,chiuzbaian2019mind,wang2021focus} \\
\bottomrule
\end{tabularx}
\label{tab:decoding}
\end{table}

\textbf{Feedback Enhancement.} Feedback enhancement in EEG signal decoding involves integrating feedback loops to elevate the user-system interaction within EEG-based platforms, for instance, BCIs. Such feedback mechanisms empower users to more adeptly adjust their brain activities, thereby boosting the system's effectiveness and ease of use. Additionally, this feedback plays a crucial role in managing particular functions, enabling and facilitating the real-time control of these operations via EEG signals \cite{qin2023application}. In our corpus, we identified six types of feedback.

\begin{itemize}
    \item Visual Feedback: The typical EEG feedback, which involves users receiving visual cues related to their brain activity or commands, such as moving objects on a screen or graphical representations of brainwave patterns (n=28). For example, video stimuli showcasing robot movements are frequently employed\cite{quiles2022cross,sugiyama2023eeg}.
    \item Auditory Feedback: Provides users with auditory signals matching their brain activity or command success, including beeps, volume changes, or complex audio information for different states or outcomes. Three studies adopted standalone audio feedback.
    \item Multimodal Feedback: Combines multiple feedback types (visual, auditory, haptic...) for an enriched user experience, potentially enhancing control efficiency. The combination of visual and auditory feedback are the preferred option \cite{braun2019prototype}. Seven studies in our corpus involved multimodal feedback.
    \item Direct Feedback: Direct feedback corresponds to the user's immediate actions (n=22). For instance, in BCI-controlled robotic arm scenario, the arm movement is the direct feedback of user's commands \cite{wei2021control,martinez2020non}.
    \item Performance Feedback: Refers to performance metrics such as algorithm accuracy and error rates, are utilized to evaluate system control and progress during training sessions (n=18).
    \item Neurofeedback: Neurofeedback provides users with real-time information about specific brainwave patterns or mental states, allowing them to learn how to enhance their cognitive abilities by modulating these patterns or states. This is the second most prevalent feedback in our corpus (n=22).
\end{itemize}

\textbf{Artificial Intelligence Methods.} The swift advancement of AI technologies in recent years has led to the creation of more sophisticated algorithms that deliver impressive performance. DL techniques, such as deep neural networks and their variations, have shown exceptional proficiency, even surpassing ML algorithms. While traditional ML methods like linear discriminant analysis \cite{jo2022eeg} and support vector machine \cite{francis2021eeg} remain popular in EEG-based BRI studies for their reliability and robustness, there is a growing trend towards advanced DL techniques, particularly in EEG signal decoding tasks such as feature extraction and classification. For example, a combination of graph convolutional networks and gated recurrent unit networks was used for feature classification in \cite{tang2022wearable}, and a blend of long-short term memory networks with convolutional neural networks was utilized for both feature extraction and classification in \cite{chenga2023using}. In our analysis, we categorized the studies based on their usage of traditional ML, DL, or non-ML/DL approaches. Despite the advent of sophisticated AI methods, our findings (Table ~\ref{tab:decoding}) indicate a continued preference for conventional ML algorithms in the majority of the studies examined (n=42).

\subsection{Robot}
The \textit{Robot} entity characterizes the autonomous or semi-autonomous machines that are capable of performing tasks or actions on their own or offering help with a degree of programmable intelligence. In our corpus, we classified robots into eight categories, focusing on their functionality, objectives, application domains, and implementation scenarios rather than traditional design aspects like robot arms or mobile robots. This approach moves beyond superficial classifications commonly found in previous reviews, offering a deeper understanding of robots' roles and uses. 
%They almost include all of the up-to-date robot/robotic systems that are deployed in current research community of BRI. 
The entire distribution of robot usage across the reviewed papers is exhibited in Table ~\ref{tab:everything}. It's crucial to realize that in our categorization, the dimensions were not exclusively mapped with the papers. For example, one study may involve multiple types of robots (i.e., \cite{yuan2018brain,zhang2021mind}) or the single robot used may be considered to have a (partial) relation to other types (i.e., \cite{prinsen2022passive,nann2020restoring}). This recognition underscores the overlapped and interconnected nature of the dimensions (robots) and facilitates an in-depth understanding of the variability and nuance within the reviewed studies.

\textit{Industrial Robot.} Industrial robots, often employed in manufacturing settings, are engineered to execute operations like welding, painting, assembly, and product handling. Renowned for their precision, speed, and durability, these robots typically manifest as robotic arms or mobile robots, catering to the rigorous demands of industrial applications for desired task performance. This type of robot emerged as the second most utilized in our corpus (n=32).

\textit{Service Robot.} Service robots are designed to aid humans in tasks beyond manufacturing realms which can be embodied in any shapes. In personal service contexts, they facilitate household chores, provide entertainment, or serve as personal aides. Conversely, in professional settings, they find application across various sectors such as healthcare, where they assist in surgeries and patient care; agriculture, through harvesting robots; and logistics, with delivery drones, among others, showcasing their versatility and utility in both personal and professional spheres. Particularly, a wheeled robot with a robot arm was harnessed for assisting disabled individuals in \cite{du2021vision}. This robot category emerged as the most favored in our analysis, with 38 studies featuring it.

\textit{Medical Robot.} Medical robots are specifically designed for healthcare applications, encompassing surgical robots that aid in procedures, rehabilitation robots for therapeutic use, and diagnostic robots for medical assessments. Common examples include wearable exoskeletons \cite{chu2018robot} for mobility support and prosthetic devices \cite{ghosh2019continuous} for limb replacement, illustrating their critical role in enhancing patient care and medical outcomes. 27 studies were identified in this dimension.

\textit{Social Robot.} Social robots are engineered to engage with humans on an interpersonal level, possessing the ability to recognize and respond to human emotions, partake in conversations, and emulate social behaviors. Typically designed as humanoid robots, these machines are crafted to closely resemble human appearance and mannerisms, facilitating natural and intuitive interactions (n=11).

\textit{Educational Robot.} Educational robots serve as interactive teaching aids, designed to assist in learning languages, programming, mathematics, and science subjects. They are specifically engineered to be engaging and interactive, making the learning experience more effective and enjoyable for students. Similarly, it is often engineered as humanoid robots to mimic human teacher actions (n=9).

\textit{Exploratory Robot.} Exploratory robots are deployed in environments deemed inaccessible or hazardous for humans, contrasting with the typical use of industrial robots. These include rovers navigating in the inaccessible places, exploring underwater regions, and searching and rescuing operations in disaster-stricken areas, showcasing their critical role in extending human reach and capabilities (n=4). Especially, a swarm robot was employed for exploratory purposes in \cite{belkacem2021cooperative}.

\textit{Autonomous Vehicle.} This innovative approach has transformed human society in recent years, introducing autonomous systems like self-driving cars, drones, and unmanned aerial vehicles (UAVs). These robots operate independently of human intervention, serving critical roles in transportation and conducting aerial surveys, marking a significant leap forward in technology and its applications (n=3). Drones were the predominant type of autonomous vehicle observed in our corpus.

\subsection{Application Contexts and Evaluation Techniques}
\subsubsection{Application Contexts}
%We thoroughly analyzed the studies in our corpus, examining their approaches concerning interactions between humans and robots. 
Based on the level of direct human control over robots, we uncovered two overarching application contexts: Human Control Concern (HCC) and Robot Design Concern (RDC). We then delineated eight more specific, but rather high-level application categories lying either on HCC or RDC contexts. Our goal is to motivate future researchers to consider the function and role of robots prior to constructing a complete BRI system under EEG. Further, we defined more specific application domains with detailed descriptions. Each study was primarily classified under either HCC or RDC, though some spanned both contexts (Table ~\ref{tab:app}).

\textbf{\textit{HCC:}} In this category, the BRI system is tailored for situations in which humans intentionally control, direct, or guide the robots to achieve specific objectives. The three inclusive sub-categories are presented below.

\textbf{\textit{RDC:}} In this category, humans intentionally controlling the robots is not the main concern; instead, the robots serve as external agents that provide assistance in carrying out specific tasks or achieving predefined goals. The five sub-categories incorporated in the RDC application context are narrated below.

Moreover, upon revisiting all studies, we identified eight application categories, with the majority present in both HCC and RDC contexts. Notably, Military Usage was uniquely associated with HCC, whereas Education was exclusively linked to RDC.

\begin{itemize}
    \item \textit{Human-centric Technology:} This category (n=28 in HCC and n=9 in RDC) investigates innovative, human-centered approaches in EEG-based BRI to enhance human interactions with robots, mutual task collaboration, and UX. A majority of the studies were research-oriented introducing new tools or methodologies to advance the field. For instance, several studies harnessed AR for enhanced robotic control \cite{chen2020combination,si2018towards}.
    \item \textit{Assistance:} Aims to leverage robotics in aiding individuals with disabilities or motor impairments, facilitating physical tasks like walking or navigation and enabling control over robots for task execution without physical movement, i.e., steering a mobile robot using solely brain signals \cite{rashid2020investigating}. 28 studies were identified with this category linked with HCC while 5 with RDC.
    \item \textit{Healthcare:} Focuses on employing advanced robotics, such as exoskeletons or prostheses, for rehabilitation and therapeutic purposes in patients with mobility disorder issues, as well as supporting individual mental recovery and meditation. In this category, 16 studies were identified under HCC and only three under RDC.
    \item \textit{Education} Targets cognitive development and skill improvement through BRI, with specific emphasis on language acquisition of gaining the ability to understand and communicate in a non-native language; and self-directed learning of obtaining knowledge or skills with self-supervision. This category was merely associated with RDC (n=5). 
    \item \textit{Entertainment:} Summarizes research where BRI systems are designed or evaluated with relaxation, fun, and leisure activities, utilizing EEG technology. One study was found in each of HCC and RDC.
    \item \textit{Safety and Security:} Investigates strategies for enhancing safety in shared human-robot environments and ensuring secure operations to boost productivity. Likewise, only one study was identified both in HCC and RDC.

\begin{landscape}
\begin{table}[!tb]
\caption{\scriptsize Application contexts with category and domain information in our corpus.}
\footnotesize\sffamily\centering%
\begin{tabularx}{\linewidth}{>{\hsize=0.4\hsize}X >{\hsize=0.7\hsize}X >{\hsize=0.6\hsize}X >{\hsize=1.8\hsize}X >{\hsize=0.45\hsize}X >{\hsize=1.15\hsize}X}
\specialrule{.2em}{.1em}{.1em}
\toprule
\textbf{Context} & \textbf{Category} & \textbf{Domain} & \textbf{Description} & \textbf{Papers (\%)} & \textbf{References} \\
\midrule
\textbf{Human Control Concern (HCC)} & Human-centric Technology & Usability and user experience & Exploring better UX for effective interactions & 7 (8\%) & \cite{fang2022brain,kim2021affect,martinez2020non,kuffuor2018brain,yuan2018brain,memar2019objective,iwane2019inferring} \\
 &  & Research and innovation & Exploring new advancements for research needs & 18 (20.7\%) & \cite{ehrlich2019feasibility,sugiyama2023eeg,kilmarx2018sequence,cao2018tactile,chen2021research,liu2021enhanced,chen2020combination,zhao2020multiple,hernandez2019deep,ak2022motor,guo2020nao,yuan2018brain,jiang2018brain,si2018towards,pawus2022application,wei2021control,cervantes2023cognidron,lu2020tractor} \\
 &  & Mutual collaboration & Collaborating with robots to enhance task performance by control & 3 (3.4\%) & \cite{lyu2022coordinating,liu2021enhanced,memar2018eeg} \\
\addlinespace
& Assistance & Physical movement & Aiding people with motor disorders for physical movements & 19 (21.8\%) &\cite{farmaki2022application,zhang2021mind,du2021vision,abougarair2021implementation,magee2021system,wang2018continuous,li2019hybrid,chiuzbaian2019mind,ghosh2019continuous,mondini2020continuous,rahul2019eeg,ali2021eeg,francis2021eeg,bahman2019robot,li2023act,long2019eeg,wang2018study,tang2022wearable,wei2021control} \\
 &  & Communication & Motionless control from people for intended actions & 7 (8\%) & \cite{boonarchatong2023green,shao2020eeg,rashid2020investigating,korovesis2019robot,karunasena2021single,quiles2022cross,bahman2019robot} \\
\addlinespace
& Healthcare & Rehabilitation & Rehabilitation for people with mobility impairment & 15 (17.2\%) &\cite{wu2022deepbrain,chu2018robot,ai2018cooperative,li2022eeg,roy2022classification,jo2022eeg,nann2020restoring,xu2018eeg,sanguantrakul2020development,araujo2021development,gordleeva2020real,chhabra2020bci,dissanayake2022eeg,wang2018study,tang2022wearable} \\
 &  & Therapy & Therapeutic process & 1 (1.1\%) & \cite{roshdy2021machine} \\
\addlinespace
& Entertainment &  & Engagement in entertaining activities & 1 (1.1\%) &\cite{cervantes2023cognidron} \\
\addlinespace
& Safety and Security & & Exploration for safe and secure needs & 1 (1.1\%) &\cite{lu2020tractor} \\
\addlinespace
& Social Interaction &  & Robots as companions for social communications & 1 (1.1\%) &\cite{toichoa2021emotion} \\
\addlinespace
& Military Usage &  & Exploration intended for military cases & 1 (1.1\%) &\cite{belkacem2021cooperative} \\
\midrule
\textbf{Robot Design Concern (RDC)} & Human-centric Technology  &  Usability and user experience & Exploring better UX for effective interactions & 2 (2.3\%) & \cite{aldini2019effect,chenga2023using} \\
 &  & Research and innovation & Exploring new advancements for research needs & 5 (5.7\%) & \cite{richter2023eeg,penaloza2018android,lu2022online,kompatsiari2018,ogino2018eeg} \\
 &  & Mutual collaboration & Collaborating with robots to enhance task performance & 2 (2.3\%) & \cite{aldini2022detection,aldini2019effect} \\
\addlinespace
& Assistance & Physical movement & Aiding people with motor disorders for physical movements & 3 (3.4\%) &\cite{yu2019design,aznan2019using,yoon2021effect} \\
 &  & Communication & Aiding motionless people for intended actions & 2 (2.3\%) & \cite{braun2019prototype,wang2021focus} \\
\addlinespace
& Healthcare & Rehabilitation & Rehabilitation for people with mobility impairment & 1 (1.1\%) &\cite{chenga2023using} \\
 &  & Mental recovery & Supports mental recovery & 1 (1.1\%) &\cite{yoon2021effect} \\
 &  & Meditation & Help with mindfulness and meditation process & 1 (1.1\%) & \cite{yoon2021effect} \\
\addlinespace
& Education & Language learning & Supports exclusively in language learning & 2 (2.3\%) &\cite{prinsen2022passive,alimardani2022robotL} \\
 &  & Self-learning & Support self-learning for skill enrichment & 3 (3.4\%) & \cite{wang2021focus,kar2021eeg,alimardani2020robotM} \\
\addlinespace
& Entertainment &  & Engagement in entertaining activities & 1 (1.1\%) &\cite{kar2021eeg} \\
\addlinespace
& Safety and Security &  & Exploration for safe and secure needs & 1 (1.1\%) &\cite{qian2018affective} \\
\addlinespace
& Social Interaction &  & Robots as companions for social communications & 4 (4.6\%) &\cite{alimardani2020robotM,baka2019talking,chang2021eeg,staffa2022enhancing} \\
\bottomrule
\specialrule{.2em}{.1em}{.1em}
\end{tabularx}
\label{tab:app}
\end{table}
\end{landscape}

    \item \textit{Social Interaction:} Encompasses studies on communication, collaboration, and companionship with robots designed to simulate more human-like interactions, focusing on robots' ability to recognize and respond to human emotions and intentions. Mostly, humanoid robots were engaged such as Pepper robot \cite{staffa2022enhancing} or Nadine robot \cite{baka2019talking}. This category includes five studies, with four in RDC and only one in HCC.
    \item \textit{Military Usage:} Explores the application of EEG-based BRI systems in military settings, with merely one study in this area associated with HCC.
\end{itemize}

\subsubsection{Evaluation Techniques}
System evaluation is pervasively used in research papers to examine the developed systems. It was observed that nearly all the reviewed studies employed this methodology to evaluate their designed BRI systems. The evaluation methodologies within our corpus were categorized into six groups: performance metrics, user experience (UX) metrics, signal information, surveys, interviews, and not specified. Some studies employed multiple categories for system evaluation (Table ~\ref{tab:eva}).

\textit{Performance metrics:} Traditional performance metrics, as typical objective measurements in EEG-based BRI system evaluation, vary depending on the specific applications and goals. 
% Various common performance metrics were adopted in the reviewed studies to assess the effectiveness so as to help humans understand how well a BRI system is functioning. 
This category represents the predominant evaluation method employed in 54 studies (i.e., \cite{sanguantrakul2020development,penaloza2018android,yu2019design}), focusing on accuracy, response time, task completion time, and error rate as main and workload and efficiency as additional.

\textit{UX metrics:} UX metrics, usually as subjective measurements, focus on gauging UX and user satisfaction when humans interact with a product within BRI. These metrics rely on the user's opinions and feelings to evaluate aspects such as usability, engagement, and perception of the system's quality. In our corpus, we spotted numerous UX metrics exploited in designing and examining the BRI systems, such as emotional states (happiness, fatigue, sorrow, excitement), usability (system usability score), and cognitive workload (NASA task load index). 16 studies embraced UX metrics for systematic evaluation (i.e., \cite{toichoa2021emotion,staffa2022enhancing,alimardani2020robotM}). 

\textit{Signal information:} This category constitutes the most preferred evaluation methodology in our corpus. Many reviewed studies chose the information gleaned from brain signals employed within their BRI systems for assessment, which in the majority of the cases, were EEG signals. This signal information encompasses various elements, including extracted features and other pertinent data. This category of evaluation methods ranked as the second most preferred approach among the studies (n=20). For instance, in \cite{chu2018robot}, information regarding steady-state visually evoked potential (a type of EEG) was extracted for evaluation, while \cite{ehrlich2019feasibility} used error-related potential (observed through EEG) for assessing their designed BRI systems.

\section{Analysis of The Interaction Entity}
\label{chap:interaction}
The aim of this paper is to investigate the recent research status with respect to EEG-based BRI systems especially to identify the interaction between brains and robots. Thus, the \textit{Interaction} entity becomes the pivotal component in our BRI system model. As aforementioned, we established theoretical innovation regarding the communication between the brain and the robot under EEG. The relevant information pertaining to the interaction mode was based on how the brain and the robot were interconnected, mutually influenced, and provided reciprocal feedback. In most instances, robots were linked with BCIs, which have previously shown high efficiency in seamless control. BCIs act as a crucial intermediary, facilitating connections between humans and external devices. However, we noted that some studies either did not employ BCIs for interaction or did not specify their involvement. In this chapter, we address the exhaustive details behind this entity, describing the definitions of the four dimensions (Pure BCI, BCI+Agents, Proactive Control, and Task-oriented HRI) with two derived sub-dimensions generated from our corpus. We illustrate their application scenarios with the practical deployment in the BRI system, featuring a particular emphasis of the control part. The overall distribution of the relevance of these dimensions within our corpus is displayed in Table ~\ref{tab:everything} while the details of sub-dimensions mapping the specific references and number of papers enumerated are presented in Table ~\ref{tab:interaction}. To note, the four dimensions with the two sub-dimensions are not exclusively mapped with any individual papers. Rather, one paper may traverse multiple dimensions, reflecting the nuanced and multifaceted nature of the interactions examined. For instance, a single paper could contribute to both Pure BCI and Proactive Control (i.e., \cite{chu2018robot,kilmarx2018sequence,cao2018tactile}), showcasing the overlap and integration of these dimensions in one study. We envision this curated knowledge being capable of benefiting future HRI research with efficient communication design. This chapter is crafted particularly for interpreting \textbf{RQ2} and \textbf{RQ3}.

\begin{table}[!t]
\caption{The evaluation methods identified in our corpus.}
\footnotesize\sffamily\centering%
\begin{tabularx}{\linewidth}{>{\hsize=0.5\hsize}X >{\hsize=0.7\hsize\raggedright\arraybackslash}X >{\hsize=0.5\hsize\centering\arraybackslash}X >{\hsize=2.0\hsize\raggedright\arraybackslash}X}
\specialrule{.2em}{.1em}{.1em}
\toprule
\textbf{} & \textbf{Metrics} & \textbf{Papers (\%)} & \textbf{References} \\
\midrule
\textbf{Evaluation Methods} & Performance metrics & 54 (62.1\%) & \cite{belkacem2021cooperative,memar2018eeg,staffa2022enhancing,iwane2019inferring,kompatsiari2018,long2019eeg,si2018towards,chenga2023using,tang2022wearable,pawus2022application,kar2021eeg,chu2018robot,kilmarx2018sequence,cao2018tactile,chen2021research,jo2022eeg,aldini2022detection,lu2020tractor,zhang2021mind,du2021vision,magee2021system,roshdy2021machine,aznan2019using,sanguantrakul2020development,penaloza2018android,yu2019design,bahman2019robot,rashid2020investigating,zhao2020multiple,guo2020nao,nann2020restoring,gordleeva2020real,korovesis2019robot,karunasena2021single,kuffuor2018brain,jiang2018brain,chen2020combination,wang2018continuous,quiles2022cross,mondini2020continuous,wei2021control,araujo2021development,rahul2019eeg,francis2021eeg,shao2020eeg,xu2018eeg,boonarchatong2023green,prinsen2022passive,li2023act,qian2018affective,farmaki2022application,lyu2022coordinating,wu2022deepbrain,sugiyama2023eeg} \\
\addlinespace
& UX metrics & 16 (18.4\%) & \cite{baka2019talking,memar2018eeg,chang2021eeg,toichoa2021emotion,staffa2022enhancing,alimardani2020robotM,chenga2023using,ogino2018eeg,cervantes2023cognidron,martinez2020non,nann2020restoring,li2023act,chhabra2020bci,fang2022brain,lyu2022coordinating,richter2023eeg} \\
\addlinespace
& Signal information & 20 (23\%) & \cite{ehrlich2019feasibility,dissanayake2022eeg,aldini2019effect,wang2021focus,lu2022online,alimardani2022robotL,chenga2023using,chu2018robot,li2022eeg,roy2022classification,aldini2022detection,penaloza2018android,yu2019design,li2019hybrid,chiuzbaian2019mind,karunasena2021single,yuan2018brain,ai2018cooperative,mondini2020continuous,kim2021affect} \\
% \addlinespace
% & Not mentioned & 9 (10.3\%) & \cite{yoon2021effect,wang2018study,liu2021enhanced,abougarair2021implementation,hernandez2019deep,ak2022motor,ghosh2019continuous,ali2021eeg,braun2019prototype} \\
\bottomrule
\specialrule{.2em}{.1em}{.1em}
\end{tabularx}
\label{tab:eva}
\end{table}

\subsection{Pure BCI}
In the EEG-based HRI context, BCI is yet the most adopted tool which focuses on enabling direct communication between the brain and robotic systems, allowing individuals to control external devices by brain signals. A fine-designed BRI system takes this concept a step further by integrating brain signals with robotic systems, empowering humans to not only control robots through the collected brain signals but also receive sensory feedback. We found that in the majority of the examined studies, BCI was extensively exploited to build the interaction between the brain and the robot. While certain studies exclusively employed BCI, others integrated BCI with complementary techniques. Hence, the first dimension of the \textit{Interaction} entity is Pure BCI, entailing those papers where the BCI was harnessed without other auxiliary technologies in bridging the connection. Almost 40\% of the studies exploited only BCIs for linking human brains and robots, yielding a dominant usage of robot arms. For example, \cite{kilmarx2018sequence} constructed the connection between the brain and a robot arm for manipulation rooted in a single BCI, while \cite{alimardani2022robotL} managed to create an assistive learning environment with the same mechanism as the robot. Particularly, exoskeletons are connected with a single BCI in two studies \cite{chu2018robot,ghosh2019continuous}.

\subsection{BCI+Agents}
In some of the studies, BCI was employed not in isolation but in conjunction with one or more supplementary technologies. We define the second dimension of this entity as BCI+agents, where additional technologies are seamlessly integrated with BCI to enhance control and interaction with robots. From our corpus, we identified 15 studies which employed external agents either physical devices or artificial technologies. In \cite{toichoa2021emotion}, the interactions were designed and conducted under the support of both BCIs and an emotion recognition sector by using the accumulated EEG signals. Whereas, some computer vision use cases such as object detection was probed with BCI, enhancing robotic precision in human-centered service designing \cite{du2021vision}. Some studies \cite{si2018towards,chen2020combination} acknowledged virtual reality (VR), augmented reality (AR), and mixed reality (MR) have flourished as the state-of-the-art technologies enabling immersive interactions in recent years and then applied XR (the catch-all term of these three techniques) with BCIs to initiate the advisable interaction in the developed BRI systems. In addition, \cite{abougarair2021implementation} linked BCI with an extra Arduino Uno microcontroller to build interaction while \cite{aznan2019using} leveraged an object detection architecture to establish interaction between human brains and robots.

\begin{table}[!htb]
\caption{The overview of the Interaction entity with the dimensions as well as the categories identified in our corpus.}
\footnotesize\sffamily\centering%
\begin{tabularx}{\linewidth}{>{\hsize=0.7\hsize}X >{\hsize=0.8\hsize\raggedright\arraybackslash}X >{\hsize=0.5\hsize\centering\arraybackslash}X >{\hsize=2.0\hsize\raggedright\arraybackslash}X}
\specialrule{.2em}{.1em}{.1em}
\toprule
\textbf{Dimension} & \textbf{Sub-dimension} & \textbf{Papers (\%)} & \textbf{References} \\
\midrule
Pure BCI & & 34(39.1\%) & \cite{memar2018eeg,dissanayake2022eeg,staffa2022enhancing,wang2021focus,iwane2019inferring,lu2022online,alimardani2022robotL,alimardani2020robotM,long2019eeg,yoon2021effect,chenga2023using,chu2018robot,kilmarx2018sequence,wang2018study,cao2018tactile,jo2022eeg,liu2021enhanced,roshdy2021machine,sanguantrakul2020development,penaloza2018android,chiuzbaian2019mind,bahman2019robot,zhao2020multiple,guo2020nao,karunasena2021single,jiang2018brain,quiles2022cross,prinsen2022passive,braun2019prototype,li2023act,farmaki2022application,chhabra2020bci,wu2022deepbrain,sugiyama2023eeg} \\
\cmidrule(lr){1-4}
BCI + Agents & & 17(19.5\%) & \cite{toichoa2021emotion,si2018towards,chen2020combination,kar2021eeg,abougarair2021implementation,zhang2021mind,du2021vision,aznan2019using,li2019hybrid,rashid2020investigating,ak2022motor,kuffuor2018brain,yuan2018brain,wang2018continuous,ai2018cooperative,fang2022brain,lyu2022coordinating} \\
\cmidrule(lr){1-4}
\textbf{Proactive Control} & \textbf{Single Signal Control (SSC)} &  47(54\%) & \cite{belkacem2021cooperative,ehrlich2019feasibility,memar2018eeg,dissanayake2022eeg,toichoa2021emotion,iwane2019inferring,memar2019objective,ogino2018eeg,cervantes2023cognidron,chu2018robot,kilmarx2018sequence,cao2018tactile,chen2021research,li2022eeg,roy2022classification,liu2021enhanced,martinez2020non,lu2020tractor,abougarair2021implementation,zhang2021mind,du2021vision,magee2021system,roshdy2021machine,sanguantrakul2020development,hernandez2019deep,chiuzbaian2019mind,bahman2019robot,rashid2020investigating,ak2022motor,zhao2020multiple,guo2020nao,karunasena2021single,kuffuor2018brain,yuan2018brain,jiang2018brain,chen2020combination,wang2018continuous,ai2018cooperative,quiles2022cross,li2023act,kim2021affect,farmaki2022application,chhabra2020bci,fang2022brain,lyu2022coordinating,wu2022deepbrain,sugiyama2023eeg} \\
& \textbf{Hybrid Signal Control (HSC)} & 9(10.3\%) & \cite{lu2022online,tang2022wearable,pawus2022application,wang2018study,jo2022eeg,li2019hybrid,gordleeva2020real,nann2020restoring,korovesis2019robot} \\
\cmidrule(lr){1-4}
Task-oriented HRI & & 17(19.5\%) & \cite{baka2019talking,dissanayake2022eeg,chang2021eeg,aldini2019effect,staffa2022enhancing,wang2021focus,kompatsiari2018,alimardani2022robotL,alimardani2020robotM,aldini2022detection,aznan2019using,penaloza2018android,yu2019design,prinsen2022passive,braun2019prototype,qian2018affective,richter2023eeg} \\
\bottomrule
\specialrule{.2em}{.1em}{.1em}
\end{tabularx}
\label{tab:interaction}
\end{table}

\subsection{Proactive Control}
The synergy between BRI systems and robot control has emerged as a breakthrough in the field of robotics. Through BRI, humans and autonomous/semi-autonomous machines are connected in unprecedented ways, enabling intuitive, and precise control of robots across a spectrum of applications. In most of the structured BRI systems (including our corpus), EEG signals are complied for directing the explicit control to robots by conveying commands. Similarly, we revealed that the majority of the studies included in our corpus had the intention to purposefully control the robots, which formed the third dimension of the \textit{Interaction} entity -- Proactive Control. The two sub-dimensions formulated according to the number of biosignals used are described in the following two sub-sections.

\subsubsection{Single Signal Control}
The main body the research studies in our corpus coincided with the condition that the EEG signal was merely engaged in flowing from brains to robots, promoting the formation of the first sub-dimension -- Single Signal Control (SSC). While a diverse array of usage relating to realistic application situations was observed, those papers conforming to SSC did not involve other physiological signals except EEG (i.e., \cite{guo2020nao,karunasena2021single,kuffuor2018brain,yuan2018brain,jiang2018brain,quiles2022cross}). In total, 39 papers were determined in accordance with SSC, where their main purpose was to achieve spontaneous robot control to meet predefined goals. Most of the studies were intended to engage merely EEG as the stimulation, to proactively control i.e., robotic arms for grasping \cite{hernandez2019deep} or mobile robots for navigating \cite{magee2021system}.

\subsubsection{Hybrid Signal Control}
While we narrated before that the ultimate corpus in this paper was determined in a fully or dominantly EEG-based context, some of the reviewed studies (n=9) leveraged multiple biosignals together with EEG for efficacious control of robots. Another sub-dimension affirmed is Hybrid Signal Control (HSC), where the compatible papers employed various types of brain signals aside from EEG to attain hybrid control with high robustness. For instance, in \cite{wang2018study}, the EOG signals were acquired together with EEG signals to blendedly supervise a mobile robot for home auxiliary. However, it's worth noting that EMG signals emerged as the most commonly employed modality in conjunction with EEG signals for HSC in our corpus \cite{pawus2022application,jo2022eeg,gordleeva2020real}. Notably, almost half of the HSC studies (n=4) concentrated on wearable robots, i.e., exoskeletons for healcare situations \cite{gordleeva2020real,nann2020restoring,tang2022wearable,jo2022eeg}.

\begin{table*}
\scriptsize
\newcommand*\rot[1]{\hbox to1em{\hss\rotatebox[origin=br]{-50}{#1}}}

\newcommand*\feature[1]{\ifcase#1 {$\times$} \or {$\pmb{\checkmark}$} \or {$\diamond$} \fi}
\newcommand*\fthr[3]{\feature#1&\feature#2&\feature#3}
\newcommand*\four[4]{\feature#1&\feature#2&\feature#3&\feature#4}
\newcommand*\five[5]{\feature#1&\feature#2&\feature#3&\feature#4&\feature#5}
\newcommand*\fsix[6]{\feature#1&\feature#2&\feature#3&\feature#4&\feature#5&\feature#6}
\newcommand*\fsvn[7]{\feature#1&\feature#2&\feature#3&\feature#4&\feature#5&\feature#6&\feature#7}
\newcommand*\feit[8]{\feature#1&\feature#2&\feature#3&\feature#4&\feature#5&\feature#6&\feature#7&\feature#8}
\makeatletter
\newcommand*\ex[6]{#1 & #2 & \four#3 & \fthr#4 & \four#5 & \fsvn#6}
\makeatother
\newcolumntype{G}{c@{}c@{}c}
\newcolumntype{L}{c@{}c@{}c@{}c}
\newcolumntype{H}{c@{}c@{}c@{}c@{}c@{}c}
\newcolumntype{J}{c@{}c@{}c@{}c@{}c@{}c@{}c}
\newcolumntype{K}{c@{}c@{}c@{}c@{}c@{}c@{}c@{}c}

\caption{\scriptsize The overview of the BRI model with (sub)dimensions identified in our corpus. {$\pmb{\checkmark}$}: completely relevant; {$\times$}: not relevant or did not mentioned ; {$\diamond$}: relevant either indirectly, implicitly, partially, or mentioned but lacked detailed descriptions. Cont. next page.}
\label{tab:everything}
\begin{tabular}{@{}ll !{\kern1.5em} L !{\kern1.5em} G !{\kern1.5em} L !{\kern2em} J@{}}
\specialrule{.2em}{.1em}{.1em}
\toprule
\multicolumn{2}{l}{LITERATURE} &
\multicolumn{4}{c}{\hspace*{-3.7cm}\parbox{4cm}{\centering \textbf{BRAIN} \\ Signal Acquisition}} & 
\multicolumn{3}{c}{\hspace*{-1.8cm}\parbox{3cm}{\centering \textbf{BRAIN} \\ Signal Decoding}} & 
\multicolumn{4}{c}{\hspace*{-0.8cm}\textbf{INTERACTION}} & 
\multicolumn{7}{c}{\hspace*{-0.8cm}\textbf{ROBOT}} \\
\midrule
\midrule
&
& \rot{EEG Paradigm}
& \rot{Acquisition Device}
& \rot{Sensor Locations}
& \rot{Numbers of Electrodes}
& \rot{Real-time Requirements}
& \rot{Feedback Enhancement}
& \rot{AI Methods}
& \rot{\textbf{Pure BCI}}
& \rot{\textbf{Auxiliary BCI}}
& \rot{\textbf{Proactive Control}}
& \rot{\textbf{Task-oriented HRI}}
& \rot{Industrial Robot}
& \rot{Service Robot}
& \rot{Medical Robot}
& \rot{Social Robot}
& \rot{Educational Robot}
& \rot{Exploratory Robot}
& \rot{Autonomous Vehicle}
\\
\midrule
\ex{\cite{qian2018affective}}                            {Qian et al 2018}                  {1121} {111} {0001} {1000000} \\
\ex{\cite{ogino2018eeg}}                            {Ogino et al 2018}                      {1111} {111} {0010} {1000000} \\
\ex{\cite{chu2018robot}}                                 {Chu et al 2018}                    {1111} {111} {1010} {0110000} \\
\ex{\cite{kilmarx2018sequence}}                         {Kilmarx et al 2018}                 {1121} {110} {1010} {1000000} \\
\ex{\cite{wang2018study}}                                 {Wang et al 2018}                  {1111} {112} {1010} {0120000} \\
\ex{\cite{cao2018tactile}}                                  {Cao et al 2018}                 {1111} {111} {0000} {0120000} \\
\ex{\cite{penaloza2018android}}                      {Penaloza et al 2018}                   {1111} {111} {1001} {0100000} \\
\ex{\cite{xu2018eeg}}                                    {Xu et al 2018}                     {1211} {111} {0010} {0110000} \\
\ex{\cite{kompatsiari2018}}                              {Kompatsiari et al 2018}            {1111} {110} {0001} {0001000} \\
\ex{\cite{si2018towards}}                               {Si-Mohammed et al 2018}             {1111} {011} {1111} {1000000} \\
\ex{\cite{kuffuor2018brain}}                      {Kuffuor et al 2018}                       {1111} {111} {0110} {1000000} \\
\ex{\cite{yuan2018brain}}                          {Yuan et al 2018}                         {1111} {211} {0110} {1000010} \\
\ex{\cite{jiang2018brain}}                        {Jiang et al 2018}                         {1211} {211} {1010} {0001000} \\
\ex{\cite{wang2018continuous}}                    {Wang et al 2018}                          {1111} {111} {0110} {1000000} \\
\ex{\cite{ai2018cooperative}}                        {Ai et al 2018}                         {1111} {210} {0110} {0210000} \\
\ex{\cite{memar2018eeg}}                               {Memar et al 2018}                    {1120} {110} {1010} {1000000} \\
\ex{\cite{hernandez2019deep}}                        {Hernandez et al 2019}                  {1111} {112} {0010} {1000000} \\
\ex{\cite{yu2019design}}                              {Yu et al 2019}                        {1111} {210} {0001} {1100000} \\
\ex{\cite{li2019hybrid}}                              {Li et al 2019}                        {1111} {111} {0110} {0110100} \\
\ex{\cite{chiuzbaian2019mind}}                      {Chiuzbaian et al 2019}                  {1111} {112} {1010} {0100001} \\
\ex{\cite{bahman2019robot}}                           {Bahman et al 2019}                    {1111} {110} {1010} {0100000} \\
\ex{\cite{aznan2019using}}                              {Aznan et al 2019}                   {1111} {111} {0101} {0100000} \\
\ex{\cite{ghosh2019continuous}}                         {Ghosh et al 2019}                   {1111} {111} {1010} {0120000} \\
\ex{\cite{baka2019talking}}                             {Baka et al 2019}                   {1111} {112} {0001} {0001000} \\
\ex{\cite{ehrlich2019feasibility}}                      {Ehrlich et al 2019}                {1111} {111} {2010} {0002100} \\
\ex{\cite{braun2019prototype}}                             {Braun et al 2019}               {1111} {111} {1002} {0010000} \\
\ex{\cite{aldini2019effect}}                            {Aldini et al 2019}                 {1111} {110} {0011} {1000000} \\
\ex{\cite{rahul2019eeg}}                              {Rahul et al 2019}                     {1111} {111} {0010} {0100000} \\
\ex{\cite{long2019eeg}}                                  {Long et al 2019}                  {1111} {111} {1111} {0110000} \\
\ex{\cite{iwane2019inferring}}                              {Iwane et al 2019}              {1011} {111} {1010} {1000000} \\
\ex{\cite{memar2019objective}}                              {Memar et al 2019}              {1001} {011} {0021} {1000000} \\
\ex{\cite{korovesis2019robot}}                       {Korovesis et al 2019}                  {1111} {111} {0010} {0100000} \\
\ex{\cite{chhabra2020bci}}                               {Chhabra et al 2020}               {1111} {111} {1010} {0210000} \\
\ex{\cite{martinez2020non}}                               {Martinez et al 2020}              {1121} {110} {1010} {0100000} \\
\ex{\cite{alimardani2020robotM}}                           {Alimardani et al 2020}          {1121} {010} {1001} {0201200} \\
\ex{\cite{rashid2020investigating}}                  {Rashid et al 2020}                     {1111} {111} {0110} {1000000} \\
\ex{\cite{zhao2020multiple}}                          {Zhao et al 2020}                      {1010} {111} {1010} {1000000} \\
\ex{\cite{guo2020nao}}                                {Guo et al 2020}                       {1111} {111} {1010} {0001000} \\
\ex{\cite{gordleeva2020real}}                       {Gordleeva et al 2020}                   {1121} {111} {0010} {0210000} \\
\ex{\cite{nann2020restoring}}                          {Nann et al 2020}                     {1111} {110} {0010} {0210000} \\
\ex{\cite{chen2020combination}}                    {Chen et al 2020}                         {1111} {111} {0110} {1000020} \\
\ex{\cite{lu2020tractor}}                              {Lu et al 2020}                       {1111} {111} {0010} {0100000} \\
\ex{\cite{sanguantrakul2020development}}        {Sanguantrakul et al 2020}                   {1111} {111} {1010} {0120000} \\
\ex{\cite{mondini2020continuous}}                     {Mondini et al 2020}                   {1121} {110} {0010} {0100000} \\
\ex{\cite{shao2020eeg}}                                {Shao et al 2020}                     {1111} {211} {0010} {0100000} \\
\ex{\cite{abougarair2021implementation}}           {Abougarair et al 2021}                   {1111} {110} {0110} {1000000} \\
\ex{\cite{zhang2021mind}}                               {Zhang et al 2021}                   {1221} {111} {0110} {0110000} \\
\ex{\cite{du2021vision}}                               {Du et al 2021}                       {1111} {111} {0110} {0120000} \\
\ex{\cite{magee2021system}}                             {Magee et al 2021}                   {1111} {111} {0010} {1000000} \\
\ex{\cite{roshdy2021machine}}                          {Roshdy et al 2021}                   {1121} {111} {1010} {0001000} \\
\ex{\cite{wei2021control}}                              {Wei et al 2021}                     {1211} {111} {0010} {0120000} \\
\ex{\cite{araujo2021development}}                    {Araujo et al 2021}                     {1111} {111} {0010} {0210000} \\
\ex{\cite{ali2021eeg}}                                {Ali et al 2021}                       {1121} {110} {0010} {0100100} \\
\ex{\cite{francis2021eeg}}                          {Francis et al 2021}                     {1001} {111} {0010} {0120020} \\
\midrule
\specialrule{.2em}{.1em}{.1em}
\end{tabular}
\end{table*}

\begin{table*}
\scriptsize
\newcommand*\rot[1]{\hbox to1em{\hss\rotatebox[origin=br]{-50}{#1}}}

\newcommand*\feature[1]{\ifcase#1 {$\times$} \or {$\pmb{\checkmark}$} \or {$\diamond$} \fi}
\newcommand*\fthr[3]{\feature#1&\feature#2&\feature#3}
\newcommand*\four[4]{\feature#1&\feature#2&\feature#3&\feature#4}
\newcommand*\five[5]{\feature#1&\feature#2&\feature#3&\feature#4&\feature#5}
\newcommand*\fsix[6]{\feature#1&\feature#2&\feature#3&\feature#4&\feature#5&\feature#6}
\newcommand*\fsvn[7]{\feature#1&\feature#2&\feature#3&\feature#4&\feature#5&\feature#6&\feature#7}
\newcommand*\feit[8]{\feature#1&\feature#2&\feature#3&\feature#4&\feature#5&\feature#6&\feature#7&\feature#8}
\makeatletter
\newcommand*\ex[6]{#1 & #2 & \four#3 & \fthr#4 & \four#5 & \fsvn#6}
\makeatother
\newcolumntype{G}{c@{}c@{}c}
\newcolumntype{L}{c@{}c@{}c@{}c}
\newcolumntype{H}{c@{}c@{}c@{}c@{}c@{}c}
\newcolumntype{J}{c@{}c@{}c@{}c@{}c@{}c@{}c}
\newcolumntype{K}{c@{}c@{}c@{}c@{}c@{}c@{}c@{}c}

\caption{Cont.}
\begin{tabular}{@{}ll !{\kern1.5em} L !{\kern1.5em} G !{\kern1.5em} L !{\kern2em} J@{}}
\specialrule{.2em}{.1em}{.1em}
\toprule
\multicolumn{2}{l}{LITERATURE} &
\multicolumn{4}{c}{\hspace*{-3.7cm}\parbox{4cm}{\centering \textbf{BRAIN} \\ Signal Acquisition}} & 
\multicolumn{3}{c}{\hspace*{-1.8cm}\parbox{3cm}{\centering \textbf{BRAIN} \\ Signal Decoding}} & 
\multicolumn{4}{c}{\hspace*{-0.8cm}\textbf{INTERACTION}} & 
\multicolumn{7}{c}{\hspace*{-0.8cm}\textbf{ROBOT}} \\
\midrule
\midrule
&
& \rot{EEG Paradigm}
& \rot{Acquisition Device}
& \rot{Sensor Locations}
& \rot{Numbers of Electrodes}
& \rot{Real-time Requirements}
& \rot{Feedback Enhancement}
& \rot{AI Methods}
& \rot{\textbf{Pure BCI}}
& \rot{\textbf{Auxiliary BCI}}
& \rot{\textbf{Proactive Control}}
& \rot{\textbf{Task-oriented HRI}}
& \rot{Industrial Robot}
& \rot{Service Robot}
& \rot{Medical Robot}
& \rot{Social Robot}
& \rot{Educational Robot}
& \rot{Exploratory Robot}
& \rot{Autonomous Vehicle}
\\
\midrule
\ex{\cite{belkacem2021cooperative}}                     {Belkacem et al 2021}                {1111} {111} {1010} {0000011} \\
\ex{\cite{kim2021affect}}                                {Kim et al 2021}                   {1121} {110} {0010} {1000000} \\
\ex{\cite{chang2021eeg}}                                 {Chang et al 2021}                 {1111} {111} {0111} {0001000} \\
\ex{\cite{toichoa2021emotion}}                          {Toichoa et al 2021}                {1121} {110} {0110} {1000000} \\
\ex{\cite{wang2021focus}}                               {Wang et al 2021}                   {1111} {212} {1001} {0002100} \\
\ex{\cite{yoon2021effect}}                              {Yoon et al 2021}                    {1121} {112} {1111} {0021000} \\
\ex{\cite{liu2021enhanced}}                              {Liu et al 2021}                    {1021} {110} {1010} {2000010} \\
\ex{\cite{karunasena2021single}}                 {Karunasena et al 2021}                     {1111} {110} {1010} {2100000} \\
\ex{\cite{kar2021eeg}}                                 {Kar et al 2021}                      {1111} {111} {0100} {0000100} \\
\ex{\cite{chen2021research}}                              {Chen et al 2021}                  {1111} {111} {0010} {1020000} \\
\ex{\cite{prinsen2022passive}}                           {Prinsen et al 2022}               {1111} {110} {1001} {0002100} \\
\ex{\cite{alimardani2022robotL}}                          {Alimardani et al 2022}           {1111} {000} {1001} {0002100} \\
\ex{\cite{quiles2022cross}}                        {Quiles et al 2022}                       {1111} {011} {1010} {2100000} \\
\ex{\cite{ak2022motor}}                                {Ak et al 2022}                       {1111} {010} {0110} {1000000} \\
\ex{\cite{farmaki2022application}}                       {Farmaki et al 2022}               {1111} {111} {1010} {0120200} \\
\ex{\cite{fang2022brain}}                                   {Fang et al 2022}               {1111} {011} {0110} {1000000} \\
\ex{\cite{lyu2022coordinating}}                            {Lyu et al 2022}                 {1121} {211} {0110} {1000000} \\
\ex{\cite{wu2022deepbrain}}                               {Wu et al 2022}                   {1221} {111} {1010} {0110000} \\
\ex{\cite{dissanayake2022eeg}}                          {Dissanayake et al 2022}            {1111} {110} {1012} {0100000} \\
\ex{\cite{staffa2022enhancing}}                          {Staffa et al 2022}                {1111} {111} {1002} {0001000} \\
\ex{\cite{tang2022wearable}}                            {Tang et al 2022}                    {1111} {111} {1111} {0110000} \\
\ex{\cite{pawus2022application}}                    {Pawus et al 2022}                      {1121} {111} {0010} {1000000} \\
\ex{\cite{li2022eeg}}                                    {Li et al 2022}                     {1110} {112} {0010} {0010000} \\
\ex{\cite{roy2022classification}}                       {Roy et al 2022}                     {1011} {111} {0010} {0010000} \\
\ex{\cite{jo2022eeg}}                                  {Jo et al 2022}                       {1121} {111} {1010} {0010000} \\
\ex{\cite{aldini2022detection}}                        {Aldini et al 2022}                   {1121} {111} {0001} {1000000} \\
\ex{\cite{lu2022online}}                              {Lu et al 2022}                        {1111} {110} {0010} {1000000} \\
\ex{\cite{sugiyama2023eeg}}                             {Sugiyama et al 2023}               {1021} {110} {1010} {1000000} \\
\ex{\cite{richter2023eeg}}                              {Richter et al 2023}                {1121} {110} {0001} {0001000} \\
\ex{\cite{boonarchatong2023green}}            {Boonarchatong et al 2023}                     {1120} {011} {0010} {1000000} \\
\ex{\cite{li2023act}}                                       {Li et al 2023}                 {1111} {111} {1010} {0120000} \\
\ex{\cite{cervantes2023cognidron}}                  {Cervantes et al 2023}                   {1111} {110} {0010} {0010001} \\
\ex{\cite{chenga2023using}}                             {Chenga et al 2023}                   {1010} {011} {1111} {0201000} \\

\midrule
\specialrule{.2em}{.1em}{.1em}
\end{tabular}
\end{table*}

\subsection{Task-oriented HRI}
HRI, as the precise term describing the collaboration between human cognition and robotic capabilities has redefined industries, enhancing productivity, safety, and adaptability. As BRI technology continues to advance, the potential for more seamless and natural interactions between humans and robots expands, promising to reshape the future of automation and human-robot partnerships. A certain number of reviewed studies (n=13) were found where the primary objective of the BRI systems was not formulated to control and instruct the robots. Instead, most of them fabricated a synergistic environment where the robots were served as an external intermediaries for human (brain) interaction, fulfilling designated tasks or operations, i.e., robot navigation \cite{chang2021eeg,aznan2019using,yu2019design} and cognitive task colloboration \cite{aldini2022detection}.
Nonetheless, we found that a small number of studies belonging to Task-oriented HRI either with a full extent or an emerging fashion also corresponded to other dimensions (i.e. Pure BCI \cite{wang2021focus}, BCI+Agents \cite{aznan2019using}, and Proactive Control \cite{aldini2019effect}).

\section{Challenges and Outlook}
\label{chap:challenges}
In the following parts, we list and elaborate on the challenges faced and potential directions for future research within the realm of EEG-based BRI that warrant exploration, based on our reviewing process. From the analysis grounded by our three inclusive entities \textit{Brain}, \textit{Robot}, and \textit{Interaction}, we realize that our corpus has indicated the up-to-date technological advancements made in the domain, whereas, several unavoidable challenges that have to be contemplated still exist. We envision the following problems along with the research outlook derived would be valuable for future investigators in this field. This chapter answers \textbf{RQ4}. 

\subsection{Signal Quality and Acquisition}

\begin{enumerate}
    \item \textbf{Low Spatial Resolution}: While EEG excels in its temporal resolution, capturing rapid changes in brain activity, it offers a notably lower spatial resolution \cite{toichoa2021emotion}. This challenge becomes particularly evident when researchers aim to identify precise brain regions involved in advanced brain imaging modalities such as functional magnetic resonance imaging or positron emission tomography scans. This inherent drawback restricts our ability to precisely pinpoint the specific anatomical location of neural activity within the brain. For future direction, complementary techniques or sophisticated source localization methods are to be upgraded to enhance the spatial precision of EEG data analysis.
    
    \item \textbf{Signal Quality}: EEG signals are inherently vulnerable to noise from a variety of sources, which can significantly jeopardize signal quality \cite{ghosh2019continuous}. The presence of noise stemming from muscle activity and eye movements whether voluntary or involuntary, can contaminate EEG recordings. Additionally, even environmental factors such as electromagnetic interference or ambient electrical noise, are able to infiltrate EEG data. The future work is bound to focus on maintaining a consistently high signal-to-noise ratio is an ongoing challenge in EEG-based BRI research. 
    
    \item \textbf{Individual Variability}: Another noteworthy challenge lies in the considerable variability revealed in EEG signals across different individuals \cite{wang2018continuous,quiles2022cross}. Different human brains possess unique physiological and functional characteristics, which inevitably result in distinct EEG patterns. This individual variability can complicate the development of BCIs that work universally for others. Consequently, customization and calibration processes are ubiquitously imperative to enable BCIs to accommodate the specific neurophysiological traits of humans.
    
    \item \textbf{Interference and Artifacts}: External sources of interference, range from everyday electrical appliances to the presence of other individuals in the vicinity, pose a substantial challenge to the integrity of EEG signals \cite{ghosh2019continuous,mondini2020continuous}. These disruptions act as interfering artifacts in EEG acquisition, which can obscure the genuine neural signals of interest. For future research, detecting and mitigating these artifacts is of paramount importance in EEG-based applications.
\end{enumerate}

\subsection{BCI Development}

\begin{enumerate}
\item \textbf{Real-time Processing}: BCIs rely on the smooth and real-time processing of EEG data to facilitate timely interactions between the human brain and the robot \cite{francis2021eeg,korovesis2019robot}. As discussed in Section ~\ref{decoding}, the significance of low latency, high accuracy, high temporal resolution, and computational efficiency in EEG-based BRI cannot be ignored, particularly with the latter two factors only evident in several studies. BCIs are increasingly employed in applications that require rapid and accurate translation of neural activity into actionable commands to be conveyed to environmental robots. To deliver a seamless and intuitive experience, the delay between the human brain signal generation and the subsequent response of the BCI-controlled agents must be minimized. Achieving this low latency processing with high accuracy necessitates a robust pipeline that can swiftly acquire and decode EEG signals. Crucially, improvements in high temporal resolution through increased sampling rates, coupled with enhanced computational efficiency are instrumental in pushing the boundaries of real-time EEG data analysis. 

\item \textbf{Training and Adaptation}: Realizing the full promise of BCIs hinges on their ability to be user-friendly and adaptive \cite{memar2019objective,richter2023eeg}. Currently, many BCIs demand humans to undergo extensive training to attain proficiency in operating the developed BRI systems effectively. This training process often involves repeated mental tasks or motor imagery to establish a reliable communication with robots. While humans may initially achieve proficient control, maintaining this level of performance over time can be challenging. Factors such as fatigue and "Gorilla effect \cite{feuchtner2018ownershift}" can lead to performance deterioration. Therefore, the adaptivity in BCI research is the central development of continuously learning and evolving alongside humans.

\item \textbf{Integration with Other Technologies}: The integration of BCIs with other technologies represents another existing challenge in HCI/HRI. This synergy is reliant on complex hardware and software systems, need to effortlessly interface with external tools or platforms, holding the potential to revolutionize a wide range of applications from healthcare to gaming and beyond \cite{lyu2022coordinating,ali2021eeg}. For example, even we identified few studies that embraced XR technique in our corpus, integrating XR into EEG-based BRI systems still remains challenging. By incorporating XR, BCIs can enable more immersive and intuitive experiences by translating human thoughts into actions within virtual environments or objects. For future avenue, selecting appropriate technologies with BCIs for interdisciplinary collaboration is crucial for expanding the horizons of HCI.

\item \textbf{User Feedback and Control}: BCIs are supposed to equip humans with comprehensive feedback and precise control over the BRI systems' operations \cite{chenga2023using,toichoa2021emotion}. However, achieving this goal can be particularly challenging, especially when BCIs primarily rely on neural signals. One possible approach is to develop intelligent interfaces that not only interpret biosignals but also provide real-time feedback to humans. The future research should formulate the feedback to be both informative and user-friendly, ensuring that individuals of varying technical backgrounds can effectively interact with the BCIs.

\item \textbf{Limited Information Bandwidth}: EEG signals inherently provide a relatively limited amount of information when compared with the intricate human cognition \cite{prinsen2022passive}. The human brain is a synthesis of complexity, forming the basis for thoughts, emotions, and actions. The EEG signal, as one kind of brain signals, represents a mere glimpse into this complexity, capturing electrical activity on the scalp resulting from underlying neural processes. This limited perspective can cause significant challenges when attempting to extract meaningful and precise information for the development of robust BCIs. In upcoming studies, advancing BCIs so that they can effectively harness the full spectrum of cognitive processes still remains to be an ambitious goal.
\end{enumerate}

\subsection{Safety and Ethical Concerns}

\begin{enumerate}
\item \textbf{Privacy and Security}: The emergence of signal acquisition tools has given rise to concerns regarding the privacy and security of human neural data \cite{wu2022deepbrain}. As i.e., BCIs continue to advance, the need for robust safeguards and ethical considerations becomes increasingly evident. While our study focuses on EEG-based BRI, the biosignal data used is highly personal and can potentially reveal intimate details about individuals' thoughts, emotions, and even medical conditions. In further investigation, ensuring the secure and safe handling of human biosignal data is an urgent challenge and needs to be addressed.

\item \textbf{Ethical and Legal Issues}: Similarly, the rapid growth of biometric data capture tools brings forth a multitude of ethical and legal considerations that require careful attention in BRI systems \cite{magee2021system,korovesis2019robot,belkacem2021cooperative}. The foremost concern revealed is the issue of informed consent. The tools used involve the acquisition and utilization of highly personal and private biosignal data, while each person involved must be provided with clear, comprehensive information about how their data will be used and the potential conflicts of interest. Furthermore, the concept of data ownership in signal-acquiring tools is a complex matter. Who has the rightful claim to the neural data generated by these BRI systems -- the individuals, the technology provider, the researchers, or a combination of them all? Another critical aspect is safeguarding against potential misuse of the technology. As these tools gain the potential to influence human thoughts and behaviors, there is a growing concern about unauthorized access and manipulation of personal body signal data. For future work, proper ethical guidelines need to be established to mitigate these risks, ensuring biosignals are being used for responsible purposes.

\item \textbf{Invasive vs Non-invasive BCIs}: The last challenge discovered from our corpus regarding BCI development is the choice between invasive and non-invasive BCIs \cite{si2018towards,quiles2022cross}. Invasive BCIs, which oblige the implantation of electrodes directly into the human brain present numerous ethical and safety concerns. Implanting electrodes might cause infection and tissue damage, which must be carefully pondered with the potential benefits. Additionally, regulatory oversight for invasive BCIs is rigorous which demands stringent safety measures. On the other hand, non-invasive BCIs (for EEG signals) are more accessible and safer in terms of physical risks, typically deliver lowering signal quality as mentioned before. This limitation can hinder their precision and reliability, impacting practical applications such as high-precision robotic control. Hence, selecting between invasive and non-invasive BCIs ultimately depends on the specific scenarios, with considering and equalizing the need of signal quality with the corresponding ethical and safety issues. 
\end{enumerate}

\subsection{User Acceptance and Accessibility}

\begin{enumerate}
\item \textbf{User Acceptance and Comfort}: For meticulously designed EEG-based BRI systems aiming at gaining widespread acknowledgement and adoption, it's imperative that they are inserted with high user acceptance and comfort \cite{si2018towards,bahman2019robot,bauer2008human}. However, achieving this can be a formidable challenge. EEG electrodes typically require close contact with the scalp, which can be discomforting especially when it is required to wear the electrodes for extended durations, such as in rehabilitation studies. Therefore, the generated user fatigue will not only diminish the UX but can also impact the accountability and reliability of the EEG signals obtained. To address this challenge, more concentration should be devoted in developing more ergonomic and unobtrusive EEG devices that minimize discomfort while maintaining signal quality, and promoting human-centered design that fosters greater user engagement and acceptance.

\item \textbf{Cost and Accessibility}: Precise EEG acquisition devices often come with a hefty price tag, which poses challenges both for research institutions and individuals interested in designing BRI systems \cite{prinsen2022passive,braun2019prototype,magee2021system}. The prohibitive cost can potentially limit the scope of BRI research, excluding smaller laboratories with budget constraints from contributing to the advancements of this field. Furthermore, this challenge restricts access for those who could benefit greatly from the equipment adopted, namely, people with motor disabilities seeking to enhance their communication with assistive devices or robots. As a result, there's a growing emphasis on democratizing the devices used in EEG-based BRI systems by making them more affordable and accessible.
\end{enumerate}

\subsection{Medical and Clinical Considerations}

\begin{enumerate}
\item \textbf{Clinical Validation}: As we disclosed, healthcare applications are getting more attention in the EEG-based BRI context, with the potential to revolutionize patient care and improve the quality of life \cite{rahul2019eeg}. However, realizing this potential which mainly relies on involving BCIs for clinical usage, needs to undergo rigorous testing and regulatory validation to demonstrate their efficacy and safety. This involves conducting extensive clinical trials and accumulating empirical evidence to support the authentic use in specific medical contexts. Achieving this is yet another prevailing challenge within future exploration.

\item \textbf{Long-term Reliability}: Ensuring the reliability and safety of the tested apparatuses over extended periods is a significant challenge, particularly when considering chronic medical applications where patients may rely on apparatuses for a lengthy duration spanning months or even years \cite{yoon2021effect,cervantes2023cognidron}. The longevity and sustained performance of i.e., BCIs are crucial for patients with chronic physiological conditions, as any deterioration in device performance can have serious consequences for their health. This challenge entails addressing issues related to signal quality and stability, skin interfaces, and the overall robustness of the BRI system, etc in upcoming related research endeavors.
\end{enumerate}

\section{Discussion}
\label{discussion}
%(1) current research landscape, (2) crucial techniques, (3) potential challenges, and (4) prospective research directions for the future development of EEG-based BRI systems

% In a nutshell, the aim of this paper is to provide a comprehensive research landscape and then light up future research directions and practice on EEG-based BRI systems from the reviewed 87 papers during 2018 -- 2023. 
In this section, we discuss key findings and formulate takeaway messages for the HRI community aligned with our BRI system model. 
% The whole corpus inclined with the findings identified are exhaustively depicted in Table ~\ref{tab:everything}.

\subsection{Current Research Landscape in EEG-Based BRI}
As we have shown in this paper, current researcher is increasingly exploring novel applications for brain-interacted robotics, ranging from assistive technologies for healthcare over collaboration and entertainment to education. We grouped these applications contexts into overarching categories of HCC and RDC, which underlines the two primary foci in the current research landscape. Analysis of our corpus reveals that the predominant application context for EEG-based BRI in recent years falls under HCC, specifically involving the control of robots by the human brain to achieve deliberate goals. Notably, scenarios where applications were aligned with human-centric technologies are particularly favored both in HCC and RDC. Healthcare and assistance applications, where robots are manipulated to achieve specific objectives, are mainly attributed to HCC. In contrast, applications for socializing wherein robots are selected and designed without clear indications of spontaneous control, instead offering companionship are mostly correlated with RDC. As for the evaluation methods, most of the reviewed papers (> 60\%) employed various performance metrics, such as accuracy, response time, task completion time, and error rate. This indicates that conventional metrics for evaluating task success remain widely adopted. The remaining two categories almost equally cover UX metrics and signal information. This likely reflects the fundamental goals of the research within the field of HCI/HRI, in which researchers often prioritize objective measures to quantitatively assess the performance and efficiency of the systems. The nearly equal distribution of these two metrics may suggest a growing importance of users' subjective experience alongside objective performance measures. UX metrics include subjective assessments, such as user satisfaction, perceived ease of use, and overall user experience, which leads to a more holistic understanding of EEG-based BRI.

Another line of current research in this field is dedicated to signal acquisition and decoding included in the \textit{Brain} entity. As for the signal acquisition, over 80\% of the papers we reviewed focused on three primary paradigms: task-based, MI, and SSVEP, known for their proven efficiency and usability in prior work. Regarding the apparatus, the Emotiv EPOC emerged as the preferred choice, a trend likely to continue until the next generation of advanced devices appear. Sensor location choices indicated a preference for targeting specific brain regions for optimal system performance, with over half of the studies not utilizing all four brain regions for EEG sensor placement. Notably, in determing the number of electrodes, the majority opted for fewer than 32, challenging the conventional wisdom that 64 electrodes are optimal. In the decoding phase, over 40\% of the studies emphasized the importance of a seamless feedback mechanism for real-time EEG decoding, enhancing both user control and experience. Whereas, we discovered that computational efficiency remains to be a challenge, possibly requiring more specialized hardware and other resources. For feedback enhancement approaches, visual feedback was the most popular representative for its convenience and intuitiveness. Notably, neurofeedback was featured in about a quarter of the studies (same to direct feedback), where we speculate that, in EEG-based environments, the profound and precise neural signals were collected and deemed to provide effective feedback. In feature extraction and classification, traditional ML algorithms were used in nearly half of the studies, owing to their proven performance in signal processing before  before the advent of the review papers (2018). Also, the advanced DL methods were in a period of rapid and prosperous development during the period of the examined studies (2018-2023). Hence, many studies incorporated various DL algorithms in signal decoding. In comparison, only several studies employed either ML or DL methods.

In relation to the \textit{Robot} entity, we have identified seven dimensions, correspond to seven types of robots according to their functionality, contextuality, and applicability, surpassing the scope of previous categorization efforts \cite{mao2017progress,aljalal2020comprehensive}. We found that service robots were most commonly used, closely followed by service robots, likely because many studies focused on using service robots, employed for human-centred services like object delivery or indoor cleaning. The prevalence of industrial robots may stem from numerous studies leveraging neural signals for precise tasks, like operating a robot arm for grasping items or guiding a mobile robot to a specific spot. Medical robots, often wearables, represent another significant dimension, aimed at aiding recovery in individuals with physical disabilities. In addition, the growing interest in social and educational robots reflects the expanding field of social robotics over the past years, with social robots taking diverse forms from humanoid \cite{staffa2022enhancing} to other designs \cite{wang2021focus}, facilitating interactive functions.

We uniquely dissected the essence of the interaction between human brains and robots, formulating the \textit{Interaction} entity which serves as the bridge facilitating mutual communication between the \textit{Brain} and \textit{Robot} entities. This is the most innovative part of our paper compared to prior review papers. Within this entity, we have categorized the design strategy into four dimensions, where Proactive Control emerged as the predominant interaction mode in our corpus (64.3\%). This might because in prevalent BRI scenarios, the driving objectives are to control and manipulate robots for desired outcomes. Of which, two more sub-dimensions are SSC and HSC. SSC, which involves using EEG signals as the sole biosignal input, constitutes the largest proportion (54\%) of brain-to-robot interactions. As a comparison, only a few studies belong to HSC (10.3\%) where other types of biosignals are harnesses at the same time. This obvious divergence is likely due to the perceived complexities, potential errors, and challenges associated with managing multi-signal uncertainty in HSC cases. The second most favoured interaction mode is Pure BCI (39.1 \%), which is twice the number of BCI+Agents (19.5 \%). This can be attributed to the convenience and ease of control of the singular adoption of BCIs. As a fact, we found that papers spanned over one dimension were predominantly categorized into both Pure BCI and Proactive Control (mostly in SSC). This observation suggests that this combination, particularly within the SSC sub-dimension, has potentially evolved into a standard in designing interaction modes for BRI systems within the reviewed literature. Simultaneously, we have to realize that the tendency of using BCIs with external agents (XR techniques, emotion recognition, etc) for effective interaction has been affirmed through our observation. About 20\% of reviewed studies were classified as Task-oriented HRI, where robots were exploited as more of providing companionship for mutual collaboration, instead of unidirectional control. We believe the similar usage will prosper with a remarkable pace in the future because of the advancements in social robotics. 

\subsection{Future Research Directions}
Within the scope of this paper, we have identified five main direction for future research directions that coincide with aforementioned challenges: (1) signal quality and acquisition, (2) BCI development, (3) safety and ethical concerns, and (4) user acceptance and accessibility, and (5) medical and clinical considerations. Future work on signal quality and acquisition can explore dry electrodes, which are more user-friendly, require less preparation time, and often yield better signal quality over extended periods. Moreover, the development of flexible and wearable electrode arrays conform to the scalp better, providing improved contact and reducing motion artifacts. Another potential direction can focus on utilizing machine learning algorithms that improve the system's ability to decode intentions accurately and enhance the flexibility and adaptability of EEG-robot interaction systems. As for the BCI development, future work should further explore signal processing techniques, such as adaptive filtering and wavelet analysis, to enhance the extraction of relevant information from EEG signals and improve the accuracy of brain signal interpretation. With the raise of ML/DL, future research can further leverage algorithms to adapt and learn user-specific patterns to enhance the system's ability to decode complex brain signals and improve control accuracy. As we have seen from this paper, not many works have focused on addressing issues related to safety and ethical concerns, which increases a demand to explore them in the future. For example, it can include feedback and notification mechanisms that ensure users are informed about the data collection, storage, and potential use of their EEG data, since obtaining clear and informed consent is essential to respect users' privacy and autonomy. Moreover, implementation of robust data security measures, including encryption and secure storage, is crucial to protect users' sensitive EEG data from unauthorized access and potential breaches. To address issues related to user acceptance and accessibility, future work should actively engage users in the design process, gather feedback, and iteratively refine the system based on user input, to ensure that the technology aligns with users' needs, preferences, and expectations, for example, by employing user-centered design. Last but not least, as for medical and clinical considerations, future work should perform rigorous clinical validation studies to assess the effectiveness and safety of the EEG-robot interaction system in real-world medical scenarios. For example, clinical trials can provide valuable evidence of the technology's clinical utility and validate its use in healthcare contexts. Additionally, collaboration with healthcare professionals, neurologists, and clinicians should be facilitated to integrate EEG-robot interaction into medical diagnoses and treatment plans.  This can ensure that the system aligns with clinical practices and contributes meaningfully to patient care.

\subsection{Limitations}
However, it's important to acknowledge a significant limitation of the model we developed and employed. This model does not explicitly encompass all the specific and potential components within the \textit{Brain} entity. We have primarily focused on EEG signal acquisition and decoding approaches with proposed dimensions (4+3) within the \textit{Brain} entity for coverage and generalizability. However, there may be more aspects not yet explored in this paper. Additionally, the associations between HCC and RDC in application contexts often depend on the specific context; for instance, a robot may be chosen to provide feedback to humans during a route guidance task, and its action control may require brain signals to generate directives. However, introducing additional layers of complexity could render both analysis and presentation unwieldy. As for the \textit{Robot} entity, while we admit the possibility of overlooking some robots or robotic systems with specialized usage, we are confident that our coverage incorporates the entirety of robots utilized in the studies published up until our examination date (31 July 2023). Nonetheless, we acknowledge that the dimensions developed in the \textit{Interaction} entity may not be in agreement with some other HRI researchers since they might have different definitions and classifications regarding interaction modes. Simultaneously, a diverse range of modalities may be employed in conjunction with BCIs, which were collectively referred as BCI+Agents in our study, as our intention was to span over a wide range of research.

%%%%
\section{Conclusion}
\label{con}

This paper delivers two principal contributions: firstly, it offers a comprehensive overview coupled with an in-depth meta-analysis of the EEG-based BRI research landscape, serving as a guiding beacon for future researchers. Secondly, it introduces a theoretical contribution in the form of an EEG-based BRI system model, meticulously delineating the constituent entities, with a primary focus on the intricacies of interaction between human brains and robots. Researchers and practitioners in this field can leverage this model as a valuable guide and checklist during the interaction design phase of EEG-based brain-robot scenarios. Specifically, our research dissects the EEG-based BRI system into three distinct entities, with a special emphasis on the design facets regarding the interaction between human brains and robots. Our findings illuminate that a significant portion of the reviewed papers prefers proactively controlling robots when establishing this interaction. Simultaneously, others are dedicated to shaping a Task-oriented HRI environment, leveraging pure BCIs, or integrating BCIs with external physical or artificial agents, with a near-even distribution among building up such interactions. Our work aims to provide a valuable compass for future researchers, enabling them to align their investigations with the proposed dimensions and sub-dimensions. This alignment facilitates the comparison of diverse related studies, thereby laying the groundwork for a unified design pipeline in the dynamic realm of EEG-based BRI. From a practical standpoint, the gained knowledge of existing research literature according to our BRI system model empowers both researchers and hands-on practitioners to navigate the design landscape for establishing effective communication between human brains and robots. However, it is crucial to acknowledge that the field of EEG-based BRI is in a state of continuous evolution, with more novel technologies to be developed in the coming future. Nonetheless, we believe that the presented work in this paper, along with the BRI system model, offers an empirically grounded foundation that can inspire future endeavors in this domain as well as bringing more possibilities for the HRI community.

\begin{acks}
To Robert, for the bagels and explaining CMYK and color spaces.
\end{acks}

\bibliographystyle{ACM-Reference-Format}
\bibliography{sample-base}

\end{document}